\documentclass[runningheads]{llncs}

\usepackage{eccv}
\usepackage{eccvabbrv}
\usepackage{graphicx}
\usepackage{booktabs}
\usepackage{multirow}
\usepackage{algorithm}
\usepackage{algpseudocode}
\newcommand{\scoredelta}[2]{\makebox[2.8em][r]{#1}\hspace{0.06em}{\scriptsize(#2)}}
\usepackage[accsupp]{axessibility}
\usepackage{docmute}
\usepackage[hidelinks,hypertexnames=false]{hyperref}
\newcommand{\suppref}[1]{\hyperref[#1]{\emph{Supp.\ Material}}}

\begin{document}

\title{OREO: Fidelity Alignment in 3D Generation via On-the-fly Rendering-Editing Optimization}

\titlerunning{OREO: Fidelity Alignment in 3D Generation}

\author{Zhiyuan Ma\inst{1,2} \and
Wenbo Hu\inst{2}\textsuperscript{\dag} \and
Wang Zhao\inst{2} \and\\
Pengfei Wang\inst{1} \and
Ying Shan\inst{2} \and
Lei Zhang\inst{1}\textsuperscript{\dag}}

\authorrunning{Z. Ma et al.}

\institute{The Hong Kong Polytechnic University
\and
Tencent ARC Lab}

\maketitle
\let\thefootnote\relax\footnotetext{\textsuperscript{\dag}Corresponding authors}

\begin{abstract}
  Despite recent advancements in 3D generation, models often struggle to produce assets with high visual fidelity.
  To bridge this gap, we propose OREO, an alignment framework that enhances the realism of 3D generators by leveraging rich 2D diffusion priors.
  Instead of relying on static datasets, OREO establishes a dynamic optimization loop that produces on-the-fly edited renderings as 2D pseudo-targets.
  At its core, we introduce Reinforced Editing, which utilizes a 2D model to refine rendered views of the 3D output, enhancing their overall visual fidelity while preserving the underlying geometry, viewpoint, and content.
  These refined views serve as high-quality supervision targets, enabling the 3D generator to learn from its own generated samples and progressively improve its visual quality.
  Experiments demonstrate that OREO effectively improves upon pre-trained baselines, producing 3D assets with enhanced visual realism.
  Our project page is at \url{https://theericma.github.io/oreo/}.
\end{abstract}


\section{Introduction}
\label{sec:intro}

\begin{figure}[!tbp]
  \centering
  \includegraphics[width=0.95\textwidth]{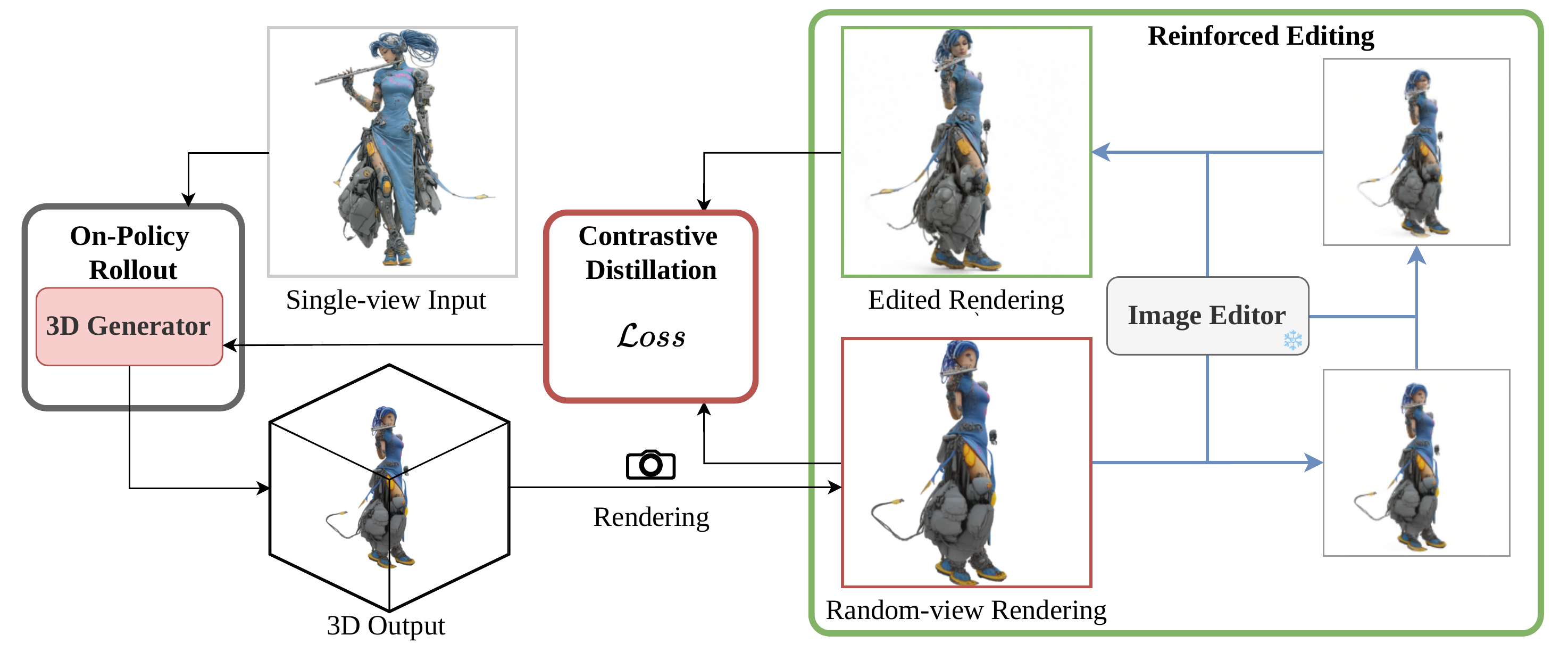}
  \caption{\textbf{Teaser.} To enhance visual fidelity of 3D generation, our framework first \textcolor{ForestGreen}{reinforces} rendered views via image editing, then distills the gap back to the training generator with a \textcolor{red}{latent contrastive} objective. Detailed workflow in Fig.~\ref{fig:pipeline}.}
  \label{fig:teaser}
  \vspace{-2mm}
\end{figure}

Recent 3D generators, such as Trellis~\cite{xiang2025structured} and the Hunyuan3D series~\cite{yang2024hunyuan3d,zhao2025hunyuan3d,hunyuan3d2025hunyuan3d,lai2025hunyuan3d2_5}, have greatly simplified the 3D content creation process.
However, despite their strong performance in geometry generation, achieving high visual fidelity remains challenging.
Generated assets often lack intricate textures and fine-grained details observed in real-world subjects.
This gap becomes more pronounced for imaginative and out-of-distribution references, where existing 3D priors provide limited appearance coverage.
This limitation largely stems from the limited scale and appearance diversity of high-quality 3D data compared with billion-scale 2D imagery.
Consequently, 3D generative models may not fully capture the visual detail needed for photorealism.

To bridge this gap, we propose to distill the rich visual priors of pre-trained 2D diffusion models into 3D generators.
These models encapsulate a deep understanding of visual fidelity and fine-grained details, far surpassing the scope of existing 3D datasets.
To this end, we introduce the On-the-fly Rendering Editing Optimization (OREO) framework, as illustrated in Fig.~\ref{fig:teaser}.
The core idea is to establish a dynamic alignment mechanism: instead of relying solely on static 3D training data, the generator improves through supervision generated on the fly.
Specifically, OREO implements a Render-Edit-Optimize loop, where the model renders views from its current output, receives corrective edits to enhance realism, and uses these refined views as supervision targets.

At the core of this loop is the Reinforced Editing algorithm, which generates edited views as supervision signals.
To ensure reliable guidance, Reinforced Editing prioritizes structure preservation, focusing on improving visual fidelity while maintaining the original spatial layout.
Inspired by recent inversion-free editing techniques, this approach mitigates the unwanted geometric deviations common in standard editing methods, helping the edited renderings better preserve the viewpoint and structure of the rendered source.
To turn these refined targets into effective supervision, we propose Contrastive Distillation, which treats the edited view as a positive and the original rendering as a negative, and distills the fidelity gap into the 3D generator through a latent-space contrastive objective.
Experiments on Trellis show that OREO produces 3D assets with improved visual fidelity.
In summary, our contributions are as follows:
\begin{enumerate}\setlength{\itemsep}{0pt}
    \item We introduce OREO, a novel alignment framework that leverages pre-trained 2D diffusion priors to enhance the fidelity of 3D generators through an on-the-fly optimization loop with dynamic 2D pseudo-targets.
    \item At its core, we propose Reinforced Editing, an image editing algorithm that improves the visual fidelity of rendered views while preserving structure, coupled with Contrastive Distillation that distills the visual changes between edited and original renderings into the 3D generator for post-training.
    \item We validate the effectiveness of OREO on Trellis, a state-of-the-art 3D generator. Results show that OREO effectively improves visual fidelity and appearance details over the pre-trained baseline.
\end{enumerate}

\section{Related Work}
\label{sec:related_work}

\noindent\textbf{Learning-based 3D Generation~\cite{ma2024scaledreamer,yang2025hybrid} and Generator Post-training.}
Recent learning-based 3D generation~\cite{xiang2025structured,zhao2025hunyuan3d,ye2025hi3dgen,li2025triposg,guo2025hyper3d,yushi2025gaussiananything,chen20253dtopia,lin2025diffsplat,li2025step1x,zhang2025bang,yang20264dvd,yang2026gencompositor,liang2026aligncvc,wang2026one2scene,liu2025mvboost,guo2026memorize,ma2025progressive} synthesizes assets feed-forward in latent spaces, but reliance on synthetic data such as Objaverse~\cite{deitke2023objaverse} caps visual fidelity and texture detail.
Photo3D~\cite{liang2026photo3d} performs offline detail enhancement on Trellis, but its edits do not explicitly preserve source-view structure.
OREO instead transfers 2D appearance priors online through structure-aligned edited renderings.

\noindent\textbf{2D Image Editing for Fidelity-Enhancing Supervision.}
Effective supervision requires edits that enhance fidelity while preserving viewpoint and structure.
Training-based image editors~\cite{brooks2023instructpix2pix,zhang2023magicbrush,sheynin2023emu,zhao2024ultraedit} inject appearance details but do not constrain viewpoint or layout, while inversion-based methods~\cite{hertz2022prompt,mokady2023null,zhang2024rfinversion} reconstruct an input before editing it.
Reinforced Editing builds on FlowEdit~\cite{couairon2024flowedit}, which couples source and target trajectories to edit images without inversion.

\noindent\textbf{2D Priors for 3D Editing and Generation.}
Although 2D priors have been explored for 3D editing and generation, existing methods do not directly meet the goal of post-training a feed-forward 3D generator.
On the one hand, methods such as Instruct-NeRF2NeRF~\cite{haque2023instruct}, DreamEditor~\cite{zhuang2023dreameditor}, GaussianEditor~\cite{wang2024gaussianeditor}, DFFSplat~\cite{koh2026diffusion}, Image Sculpting~\cite{yenphraphai2024image}, MvDrag3D~\cite{chen2024mvdrag3d}, and Omni-3DEdit~\cite{chen2026omni} edit individual instances or scenes, rather than post-training a generalizable 3D generator.
GeoDiffusion~\cite{chen2024geodiffusion} instead uses geometric conditions to control image generation for object detection data.
On the other hand, score-distillation methods such as DreamFusion~\cite{poole2022dreamfusion}, Magic3D~\cite{lin2023magic3d}, and ProlificDreamer~\cite{wang2024prolificdreamer} are optimization-based text-to-3D methods whose input setting and inference pipeline differ from those used by OREO for learning-based image-to-3D generator post-training, making them unsuitable for direct quantitative comparison with our framework.
DMD~\cite{yin2024onestep} is closer to a generator-learning paradigm and is therefore retained as a reasonable score-distillation ablation baseline; however, it still relies on implicit score-gradient supervision.
OREO uses high-fidelity, structure-aligned edited renderings as explicit 2D pseudo-targets to provide stable online supervision for 3D generator post-training.

\section{Method}
\label{sec:method}

\begin{figure}[!tbp]
  \centering
  \includegraphics[width=\textwidth]{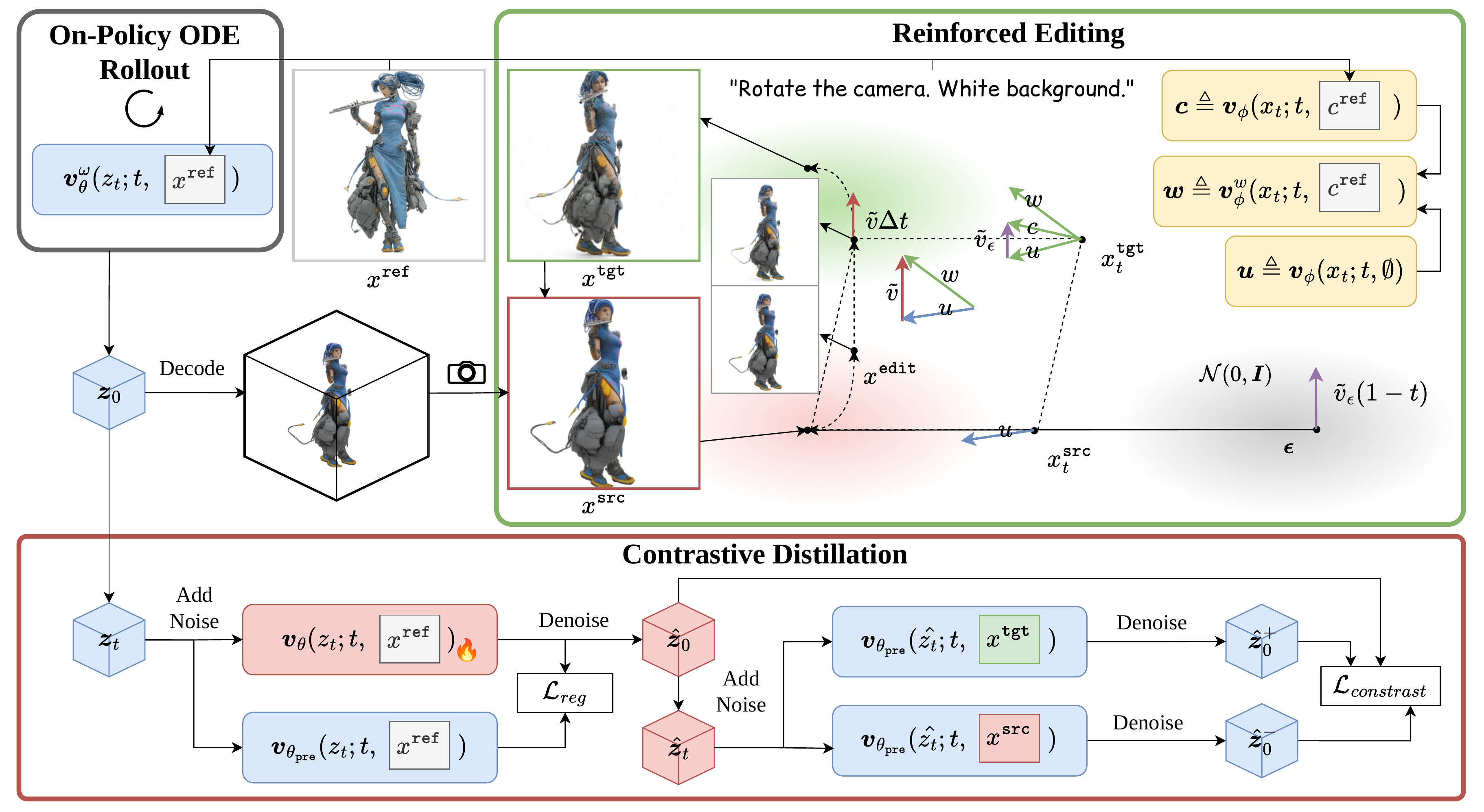}
  \caption{\textbf{Overview of OREO.} The training generator produces a 3D asset via on-policy ODE rollout that is rendered into a source view $x^{\texttt{src}}$. \textcolor{ForestGreen}{Reinforced Editing} turns $x^{\texttt{src}}$ into a high-fidelity target $x^{\texttt{tgt}}$, and \textcolor{red}{Contrastive Distillation} uses $(x^{\texttt{src}}, x^{\texttt{tgt}})$ as a negative/positive pair to supervise the generator in the latent space.}
  \label{fig:pipeline}
\end{figure}

We present OREO, an alignment framework that enhances 3D visual fidelity via on-the-fly optimization.
An overview of the full pipeline is illustrated in Fig.~\ref{fig:pipeline}.
We leverage a pre-trained 2D editing model to construct edited renderings as 2D pseudo-targets on the fly, enabling generator post-training without paired 3D ground-truth supervision.
Each iteration rolls out the generator to obtain a 3D asset and renders it from a sampled camera pose.
Building on the flow-matching and inversion-free editing formulation in Sec.~\ref{sec:preliminaries}, Reinforced Editing in Sec.~\ref{sec:reinforced_editing} enhances the view while preserving structure.
Sec.~\ref{sec:optimization} then distills this improvement into the 3D generator.

\subsection{Problem Formulation}
\label{sec:problem_formulation}
Given a reference image $x^{\texttt{ref}}$, the 3D latent generator $G_\theta$ followed by a decoder $\mathcal{D}$ produces a 3D asset $\mathcal{A} = \mathcal{D}(G_\theta(x^{\texttt{ref}}))$.
A renderer $\mathcal{P}$ then projects $\mathcal{A}$ into a 2D view $x^{\texttt{src}} = \mathcal{P}(\mathcal{A}, \pi)$ under camera pose $\pi$.
In our training framework, a pre-trained 2D editor $\mathcal{E}_\phi$ then enhances this view into a higher-fidelity target $x^{\texttt{tgt}} = \mathcal{E}_\phi(x^{\texttt{src}}, x^{\texttt{ref}})$.
We optimize the generator by minimizing this discrepancy:
\begin{equation}
\label{eq:task}
\theta^* = \arg\min_\theta \mathbb{E}_{x^{\texttt{ref}}, \pi}
\left[\mathcal{L}(x^{\texttt{src}}, x^{\texttt{tgt}})\right]
\end{equation}
where $\mathcal{L}$ is a supervision loss that encourages improved visual fidelity.

\subsection{Preliminaries}
\label{sec:preliminaries}
\textbf{Flow Matching Framework and Notation.}
Flow Matching models generation as an ODE process that transports Gaussian noise to the data distribution:
\begin{equation}
    dx_t = v(x_t; t, c) dt, \quad t: 1 \rightarrow 0
\end{equation}
where $x_t$ is the latent state at time $t$, and $c$ is the condition (e.g., text or image).
For better alignment to the condition $c$, Classifier-Free Guidance (CFG) is applied to the velocity field, using a null condition $\emptyset$ and a guidance scale $w$:
\begin{equation}
    \label{eq:cfg_guidance}
    v^{w}(x_t; t, c) = v(x_t; t, \emptyset) + w \cdot (v(x_t; t, c) - v(x_t; t, \emptyset))
\end{equation}
We estimate the clean sample $\hat{x}_0$ from the intermediate state $x_t$ in one step:
\begin{equation}
    \label{eq:single_step_estimation}
    \hat{x}_0 = x_t - t \cdot v(x_t; t, c)
\end{equation}
To distinguish modalities, we use $z$ and $\theta$ for the 3D generator $v_\theta$, where $G_\theta$ denotes its ODE sampling process and the pre-trained frozen weights are denoted $v_{\theta_{pre}}$.
Similarly, for the 2D editor $v_\phi$, we use $x$ for its latent state and $\phi$ for its parameters; $\mathcal{E}_\phi$ denotes the multi-step editing process described below.

\textbf{Inversion-Free Image Editing.}
We leverage FlowEdit~\cite{couairon2024flowedit}, an inversion-free method that modifies images by coupling source and target flow trajectories.
Starting from $x^{\texttt{src}}$, it updates $x^{\texttt{edit}} \leftarrow x^{\texttt{edit}} - \Delta t \cdot \tilde{v}$, where $\Delta t = t_i-t_{i+1}>0$ for the decreasing ODE schedule, with differential velocity:
\begin{equation}
    \label{eq:flowedit_transition}
    \tilde{v} = v^{\texttt{tgt}}(x_t^{\texttt{tgt}}; t, c^{\texttt{tgt}}) - v^{\texttt{src}}(x_t^{\texttt{src}}; t, c^{\texttt{src}})
\end{equation}
where the \textbf{source branch} and the \textbf{target branch} are constructed as
\begin{equation}
    \label{eq:flowedit_states}
    x_t^{\texttt{src}} = (1-t)x^{\texttt{src}} + t\epsilon, \quad x_t^{\texttt{tgt}} = x^{\texttt{edit}} + (x_t^{\texttt{src}} - x^{\texttt{src}}), \quad \epsilon \sim \mathcal{N}(0, I).
\end{equation}
The editing process is performed over the last $N_e$ steps of the full $N$-step ODE schedule.
These steps evolve $x^{\texttt{edit}}$ into $x^{\texttt{tgt}}$, transferring the image from $c^{\texttt{src}}$ to $c^{\texttt{tgt}}$ while preserving structure without costly inversion to noise.

\begin{algorithm}[t]
    \caption{Reinforced Editing}
    \label{alg:red}
    \begin{algorithmic}[1]
    \Require Rendered view $x^{\texttt{src}}$, Reference condition $c^{\texttt{ref}}$, Editor $\mathcal{E}_\phi$, Guidance scale $w$, Editing steps $N_e$, Total steps $N$
    \Ensure Edited rendering pseudo-target $x^{\texttt{tgt}}$
    \State Initialize $x^{\texttt{edit}} \leftarrow x^{\texttt{src}}$, Sample noise $\epsilon \sim \mathcal{N}(0, I)$
    \State Select time steps $\{t_i\}_{i=0}^{N_e}$ as the last $N_e$ steps from the full schedule of $N$ steps
    \For{$i = 0$ to $N_e-1$} \Comment{\texttt{FlowEdit: Reverse-time Euler Integration}}
        \State $t \leftarrow t_i$, $\Delta t \leftarrow t_i - t_{i+1}$
        \State $x_t^{\texttt{src}} \leftarrow (1-t)x^{\texttt{src}} + t\epsilon$
        \State $x_t^{\texttt{tgt}} \leftarrow x^{\texttt{edit}} + (x_t^{\texttt{src}} - x^{\texttt{src}})$ \Comment{\texttt{Eq.~\ref{eq:flowedit_states}}}
        
        \Statex \texttt{// Reinforced Editing (Sec.~\ref{sec:reinforced_editing})}
        \State $\tilde{v} \leftarrow v_\phi^{w}(x_t^{\texttt{tgt}}; t, c^{\texttt{ref}}) - v_\phi(x_t^{\texttt{src}}; t, \emptyset)$ \Comment{\texttt{Eq.~\ref{eq:red_diff_velocity}}}
        \State $x^{\texttt{edit}} \leftarrow x^{\texttt{edit}} - \tilde{v} \Delta t$ \Comment{\texttt{FlowEdit: Euler Step}}
    
        \Statex \texttt{// Noise Update}
        \State $\epsilon \leftarrow \epsilon - (v_\phi(x_t^{\texttt{tgt}}; t, c^{\texttt{ref}}) - v_\phi(x_t^{\texttt{tgt}}; t, \emptyset)) \cdot (1-t)$ \Comment{\texttt{Eq.~\ref{eq:red_dynamic_noise}}}
    \EndFor
    \State \Return $x^{\texttt{tgt}} \leftarrow x^{\texttt{edit}}$
    \end{algorithmic}
    \end{algorithm}

\subsection{On-the-fly 2D Pseudo-Targets via Reinforced Editing}
\label{sec:reinforced_editing}

A critical step in our pipeline is transforming the rendered view $x^{\texttt{src}}$ into an edited rendering $x^{\texttt{tgt}}$, which serves as a 2D pseudo-target for appearance supervision.
Existing image editing models~\cite{wu2025qwen} can leverage their learned priors to make rendered views more realistic, but often alter the camera perspective or object pose, as analyzed in Sec.~\ref{sec:exp_feedback}.
We propose Reinforced Editing to enable structure-preserving enhancement on top of pre-trained editing models.
Different from the original FlowEdit~\cite{couairon2024flowedit} that transfers between separate conditions $c^{\texttt{src}}$ and $c^{\texttt{tgt}}$ in Eq.~\ref{eq:flowedit_transition}, our goal is to enhance the rendered view to better resemble a novel view of $x^{\texttt{ref}}$, where no independent $c^{\texttt{src}}$ exists.
We thus guide the target branch toward $c^{\texttt{ref}} = c^{\texttt{tgt}}$ and replace the source branch with an unconditional prediction:
\begin{equation}
    \label{eq:red_diff_velocity}
    \tilde{v} = v_\phi^{w}(x_t^{\texttt{tgt}}; t, c^{\texttt{ref}}) - v_\phi(x_t^{\texttt{src}}; t, \emptyset),
\end{equation}
where $v_\phi$ is the velocity field of the pre-trained 2D editor, $w$ is the classifier-free guidance scale~\cite{ho2021classifier}, and $c^{\texttt{ref}}$ encodes both the reference image $x^{\texttt{ref}}$ and a text prompt indicating a viewpoint change, as illustrated in Fig.~\ref{fig:pipeline}.

\textbf{Noise Update.}
Random noise can weaken the editing strength~\cite{xie2026dnaedit}.
We initialize the shared noise $\epsilon$ once per rendered view, then update it at each editing step by injecting the classifier-free guidance signal:
\begin{equation}
    \label{eq:red_dynamic_noise}
    \epsilon \leftarrow \epsilon - (v_\phi(x_t^{\texttt{tgt}}; t, c^{\texttt{ref}}) - v_\phi(x_t^{\texttt{tgt}}; t, \emptyset)) \cdot (1-t)
\end{equation}
This update aligns the noise with the conditional direction to strengthen editing while preserving fine-grained structural details. Algorithm~\ref{alg:red} summarizes the complete procedure, including the noise update at each editing step.

\begin{algorithm}[t]
    \caption{Closed-loop Optimization via Dynamic 2D Pseudo-Targets}
    \label{alg:oreo}
    \begin{algorithmic}[1]
    \Require Generator $v_\theta$, frozen pretrained velocity $v_{\theta_{pre}}$, 2D editor $\mathcal{E}_\phi$, learning rate $\eta$
    \Ensure Optimized parameters $\theta^*$
    \While{not converged}
        \State Given reference image $x^{\texttt{ref}}$
        \State $z_0 = G_\theta(x^{\texttt{ref}})$ \Comment{\texttt{Rollout}: Generate clean latent}
        \State Sample camera pose $\pi$
        \State $x^{\texttt{src}} = \mathcal{P}(\mathcal{D}(z_0), \pi)$ \Comment{\texttt{Render}: Decode and project}
        \State Obtain $x^{\texttt{tgt}}$ via Algorithm~\ref{alg:red} from $x^{\texttt{src}}$ and $x^{\texttt{ref}}$ \Comment{\texttt{Reinforced Editing}}
        \State Sample $t \sim \mathcal{U}(0,1)$
        \State $z_t = (1-t)z_0 + t\epsilon,\ \epsilon \sim \mathcal{N}(0,I)$ \Comment{\texttt{Noise Injection}}
        \State Predict $\hat{z}_0$ from $z_t$ using $v_\theta$ \Comment{\texttt{Prediction}: Denoise latent, Eq.~\ref{eq:single_step_estimation}}
        \State $\hat{z}_t = (1-t)\hat{z}_0 + t\epsilon',\ \epsilon' \sim \mathcal{N}(0,I)$ \Comment{\texttt{Re-noise anchor prediction}}
        \State $z_0^{+} = \hat{z}_t - t \cdot v_{\theta_{pre}}(\hat{z}_t; t, x^{\texttt{tgt}})$ \Comment{\texttt{Detached positive}}
        \State $z_0^{-} = \hat{z}_t - t \cdot v_{\theta_{pre}}(\hat{z}_t; t, x^{\texttt{src}})$ \Comment{\texttt{Detached negative}}
        \State $\mathcal{L}_{contrast} = \| \hat{z}_0 - z_0^{+} \|^2 - \| \hat{z}_0 - z_0^{-} \|^2$ \Comment{\texttt{Latent contrastive loss}, Eq.~\ref{eq:contrast_loss}}
        \State $\mathcal{L}_{reg} = \| v_\theta(z_t) - v_{\theta_{pre}}(z_t) \|^2$ \Comment{\texttt{Regularization}}
        \State $\theta \leftarrow \theta - \eta \nabla_\theta (\mathcal{L}_{contrast} + \lambda \mathcal{L}_{reg})$ \Comment{\texttt{Update generator}}
    \EndWhile
    \end{algorithmic}
    \end{algorithm}

\subsection{Contrastive Distillation with 2D Pseudo-Targets}
\label{sec:optimization}

To instantiate the loss $\mathcal{L}$ in Eq.~\ref{eq:task} for the 3D latent generator $G_\theta$, we lift supervision from the rendered views $x^{\texttt{src}}, x^{\texttt{tgt}}$ into the latent space of the pretrained generator $v_{\theta_{pre}}$, and cast it as a contrastive objective: we treat the generator's predicted clean latent $\hat{z}_0$ as an \emph{anchor}, with the edited rendering $x^{\texttt{tgt}}$ inducing a positive latent target and the original rendering $x^{\texttt{src}}$ inducing a negative latent target. This closes the Render-Edit-Optimize loop.

\textbf{Per-Iteration Setup.}
Concretely, in each training iteration, given a reference image $x^{\texttt{ref}}$, we roll out a clean latent $z_0 = G_\theta(x^{\texttt{ref}})$ via ODE sampling.
We then sample a camera pose $\pi$ and render $x^{\texttt{src}} = \mathcal{P}(\mathcal{D}(z_0), \pi)$.
From it we obtain an edited rendering $x^{\texttt{tgt}}$ via Reinforced Editing in Algorithm~\ref{alg:red}.
Finally, we sample $t \sim \mathcal{U}(0,1)$ and perturb the latent as $z_t = (1-t)z_0 + t\epsilon$, with $\epsilon \sim \mathcal{N}(0,I)$. A single denoising step predicts the clean latent:
\begin{equation}
\hat{z}_0 = z_t - t \cdot v_\theta(z_t; t, x^{\texttt{ref}}),
\end{equation}
which serves as the anchor in our contrastive objective below.

\textbf{Latent-Space Contrastive Supervision.}
Inspired by~\cite{yu2026self}, we re-noise the anchor with fresh noise $\epsilon' \sim \mathcal{N}(0,I)$ as $\hat{z}_t = (1-t)\hat{z}_0 + t\epsilon'$, and run two passes of the frozen pretrained generator with shared $\hat{z}_t$ and different visual conditions:
\begin{equation}
\label{eq:contrast_pos_neg}
z_0^{+} = \hat{z}_t - t \cdot v_{\theta_{pre}}(\hat{z}_t; t, x^{\texttt{tgt}}),
\quad
z_0^{-} = \hat{z}_t - t \cdot v_{\theta_{pre}}(\hat{z}_t; t, x^{\texttt{src}}).
\end{equation}
Both $z_0^{+}$ and $z_0^{-}$ are detached from the computational graph and serve as the positive and the negative, respectively. Under the edited rendering $x^{\texttt{tgt}}$, the pretrained generator denoises toward the edited view, while under $x^{\texttt{src}}$ it denoises toward the un-edited rendering.
Our contrastive objective pulls the anchor $\hat{z}_0$ toward the positive $z_0^{+}$ and pushes it away from the negative $z_0^{-}$:
\begin{equation}
\label{eq:contrast_loss}
\mathcal{L}_{contrast} = \| \hat{z}_0 - z_0^{+} \|^2 - \| \hat{z}_0 - z_0^{-} \|^2.
\end{equation}
For geometric stability, we regularize the velocity against the pre-trained prior:
\begin{equation}
\mathcal{L}_{reg} = \| v_\theta(z_t) - v_{\theta_{pre}}(z_t) \|^2,
\end{equation}
giving the total objective $\mathcal{L}_{total} = \mathcal{L}_{contrast} + \lambda \mathcal{L}_{reg}$.
Algorithm~\ref{alg:oreo} summarizes the complete loop of rollout, editing, and generator updates.

\section{Experiments}
\label{sec:experiments}

\subsection{Experimental Setup}
\label{sec:exp_setup}

\textbf{Datasets.}
Existing 3D benchmarks such as Objaverse~\cite{deitke2023objaverse} subsets focus on everyday objects, where pre-trained baselines already do well and fidelity gaps stay hidden.
To cover the more complex needs of image-to-3D users, especially imaginative and diverse concept-design tasks, we curate a \textit{Conceptual Design Dataset} of 2{,}396 in-the-wild reference images under-served by existing 3D priors.
We hold out 100 references for evaluation and use the remaining 2{,}296 for training. OREO uses these images alone and requires no paired 3D ground truth.
Curation and filtering details are in \suppref{sec:dataset_construction}.
For a general-distribution reference, we additionally evaluate on Google Scanned Objects (GSO)~\cite{downs2022google}.

\textbf{Implementation Details.}
We use Trellis~\cite{xiang2025structured} as the 3D backbone and Qwen-Image-Edit~\cite{wu2025qwen} for Reinforced Editing, whose hyper-parameters are given in Sec.~\ref{sec:exp_feedback}.
Training details are provided in \suppref{sec:training_details}.

\begin{figure*}[!tbp]
    \centering
    \includegraphics[width=\textwidth]{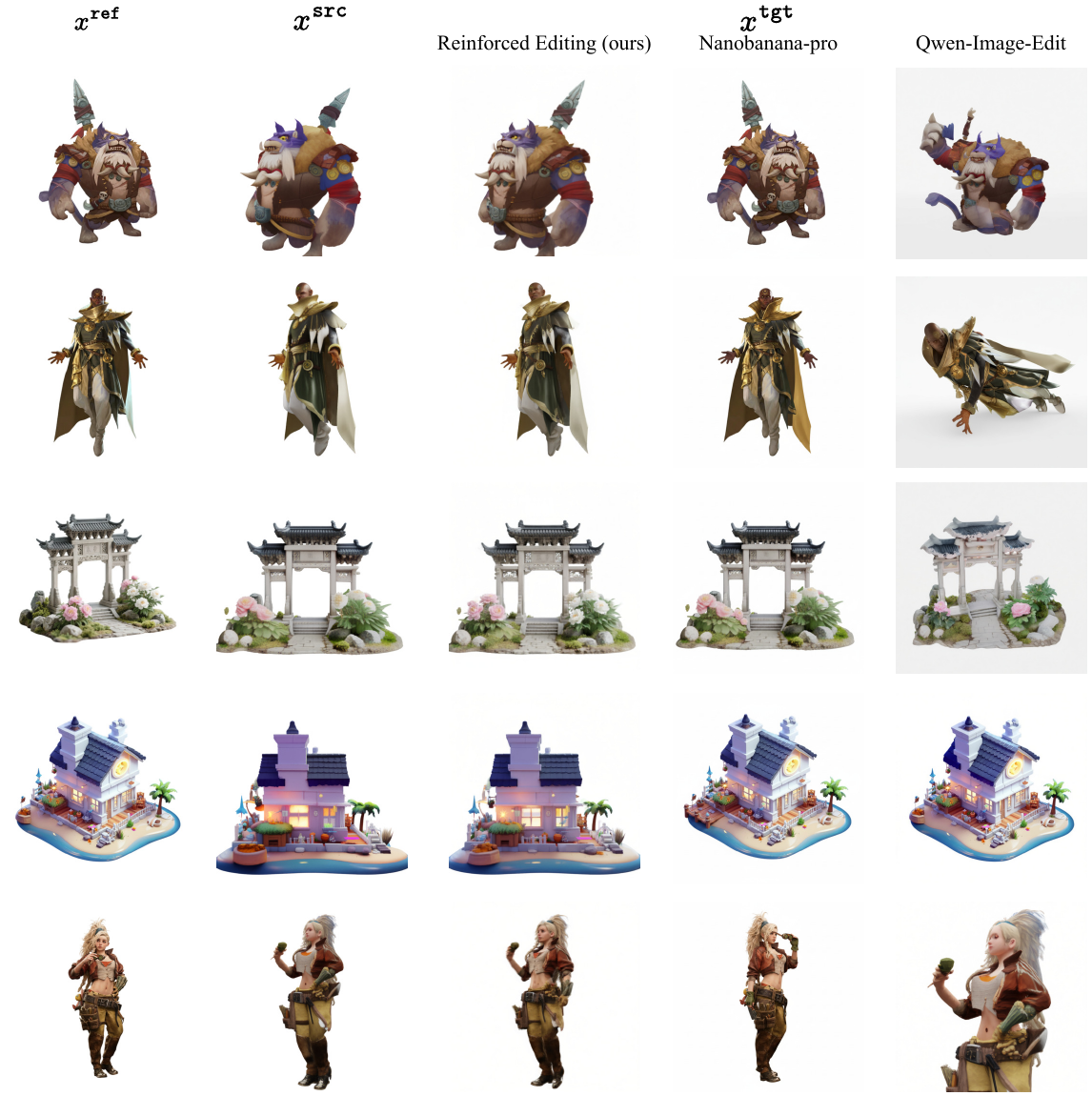}
    \caption{\textbf{Qualitative comparison of 2D feedback sources.} An ideal $x^{\texttt{tgt}}$ keeps the viewpoint of $x^{\texttt{src}}$ but lifts its fidelity toward $x^{\texttt{ref}}$; we compare Reinforced Editing (Ours), NanoBanana Pro, and Qwen-Image-Edit. Details are discussed in Sec.~\ref{sec:exp_feedback}.}
    \label{fig:edit_comparison}
  \end{figure*}

\begin{table*}[!tbp]
    \centering
    \caption{\textbf{Quantitative comparison of 2D feedback sources.} We report CLIP and DINO similarity to $x^{\texttt{ref}}$ for fidelity and to $x^{\texttt{src}}$ for content preservation, together with Mask IoU for viewpoint consistency. See Sec.~\ref{sec:exp_feedback}.}
    \label{tab:feedback_comparison}
    \setlength{\tabcolsep}{3pt}
    \resizebox{\textwidth}{!}{%
    \begin{tabular}{@{}lccccc@{}}
      \toprule
      \multirow{2}{*}{Method} & \multicolumn{2}{c}{w.r.t.\ $x^{\texttt{ref}}$} & \multicolumn{2}{c}{w.r.t.\ $x^{\texttt{src}}$} & \multirow{2}{*}{IoU $\uparrow$} \\
      \cmidrule(lr){2-3}\cmidrule(lr){4-5}
      & CLIP Sim. $\uparrow$ & DINO Sim. $\uparrow$ & CLIP Sim. $\uparrow$ & DINO Sim. $\uparrow$ & \\
      \midrule
      Unedited & 0.7613 & 0.7916 & N/A & N/A & N/A \\
      \cmidrule(lr){1-6}
      NanoBanana Pro & \textbf{0.8803} & \textbf{0.9001} & 0.8199 & 0.8404 & 0.6830 \\
      Qwen-Edit (Orig.) & 0.7753 & 0.8269 & 0.7882 & 0.8084 & 0.6374 \\
      \midrule
      Ours & \scoredelta{0.7992}{+0.0379} & \scoredelta{0.8214}{+0.0298} & \textbf{0.8833} & \textbf{0.9099} & \textbf{0.9520} \\
      \bottomrule
    \end{tabular}}
  \end{table*}

\subsection{Validating the Editing Feedback}
\label{sec:exp_feedback}

The 3D generator is supervised by edited-rendering pseudo-targets from a 2D editor, so the quality of these edits directly bounds what the 3D loop can achieve.
We therefore first evaluate the editor's feedback at the 2D level.

\textbf{Evaluation Protocol.}
For each of the 100 held-out references in the Conceptual Design Dataset, we generate a 3D asset with Trellis and render 5 views at azimuths $-90^\circ$, $-45^\circ$, $0^\circ$, $45^\circ$, and $90^\circ$ as source views $x^{\texttt{src}}$.
To evaluate the editing performance, we adopt three metrics.
To measure the multi-view fidelity improvement, we compute the CLIP-ViT-L/14~\cite{radford2021learning} and DINOv3-ViT-L/16~\cite{simeoni2025dinov3} embedding similarity of each rendered view to the reference $x^{\texttt{ref}}$, and report the mean over the 5 views before and after editing. To check content preservation, we additionally report the similarity to $x^{\texttt{src}}$.
Additionally, to quantify viewpoint and silhouette preservation, we compute Mask IoU between the edited output and $x^{\texttt{src}}$, using foreground masks extracted by U$^2$-Net~\cite{qin2020u2} as in Trellis.

\textbf{Compared Methods.}
We compare our Reinforced Editing (RE) against the original Qwen-Image-Edit~\cite{wu2025qwen} pipeline and NanoBanana Pro.
The setup for the compared methods is provided in \suppref{sec:editing_setup}.

\textbf{Qualitative Analysis.}
Fig.~\ref{fig:edit_comparison} compares the feedback sources along two requirements: improving appearance while preserving the viewpoint and content of $x^{\texttt{src}}$.
NanoBanana Pro often follows the reference image too aggressively, replacing the source viewpoint with a frontal view and altering local content such as the mage's cape.
Qwen-Image-Edit introduces larger structural changes, including pose deformation, cropping, and background drift.
In contrast, Reinforced Editing retains the source perspective and layout while adding appearance details, producing edited renderings that are better aligned with the rendered views.

\textbf{Quantitative Results.}
Table~\ref{tab:feedback_comparison} tests whether each editor provides the desired appearance-improvement direction. Similarity to $x^{\texttt{ref}}$ measures semantic and appearance alignment, while similarity to $x^{\texttt{src}}$ and Mask IoU assess whether the source-view structural conditions remain stable. The unedited row provides the baseline, and the positive $\Delta$CLIP/$\Delta$DINO values shown for RE simply confirm that editing improves its alignment with $x^{\texttt{ref}}$.
NanoBanana Pro scores higher against $x^{\texttt{ref}}$ but lower against $x^{\texttt{src}}$ and on Mask IoU, indicating that its apparent gain is entangled with changes to the rendered view. RE provides a cleaner fidelity-improvement signal: it moves the rendered view toward the reference in appearance while avoiding unwanted changes in viewpoint, silhouette, pose, and spatial layout. This makes its edited renderings more suitable as 2D pseudo-targets for post-training the 3D generator.

\setcounter{figure}{4}
\begin{figure*}[!tbp]
  \centering
  \includegraphics[width=\textwidth]{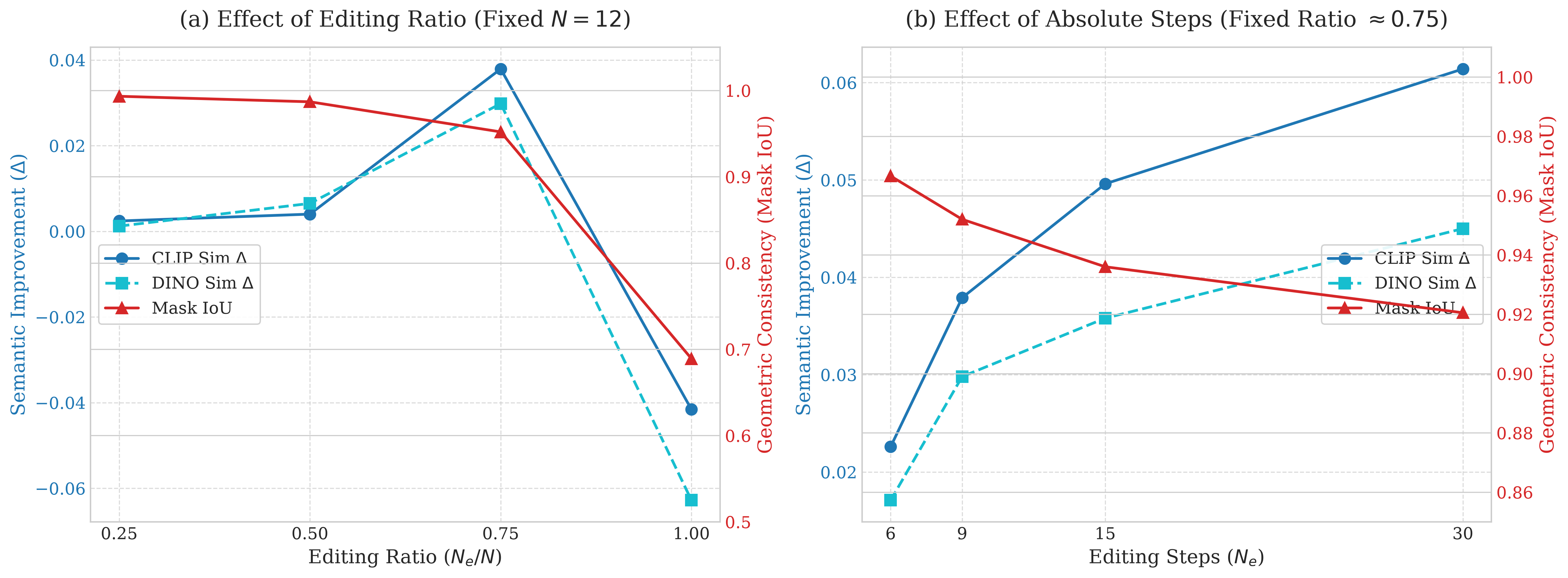}
  \captionof{figure}{Sensitivity Analysis of RE Parameters. (a) \textbf{Effect of Editing Ratio ($N_e/N$)}: A ratio of 0.75 balances semantic improvement (CLIP/DINO $\Delta$) and structural consistency (Mask IoU). (b) \textbf{Effect of Editing Steps ($N_e$)}: Under a fixed ratio ($\approx 0.75$), more steps improve CLIP/DINO similarity but reduce Mask IoU and increase cost, motivating our choice of 9 editing steps as a balanced setting.}
  \label{fig:ablation_ratio_steps}
  \vspace{2mm}
  \begin{subfigure}[b]{\textwidth}
    \centering
    \includegraphics[width=\textwidth]{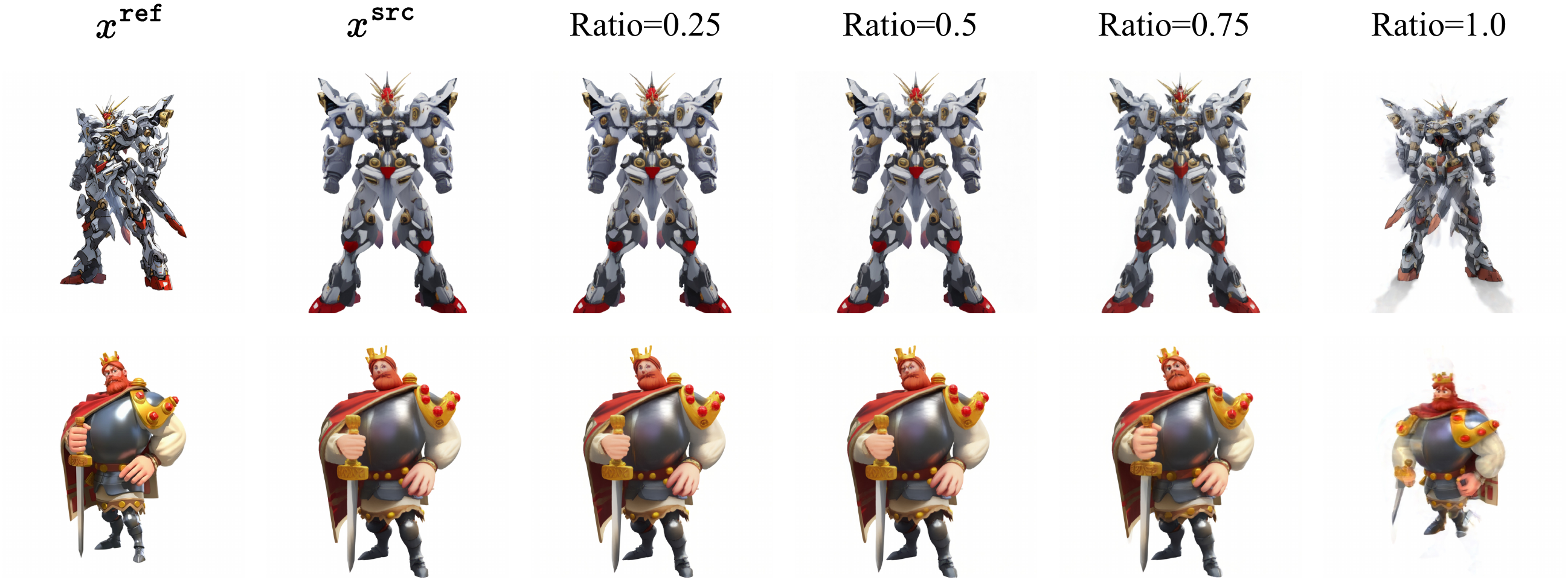}
    \caption{Effect of Editing Ratio ($N_e/N$)}
  \end{subfigure}
  \par\smallskip
  \begin{subfigure}[b]{\textwidth}
    \centering
    \includegraphics[width=\textwidth]{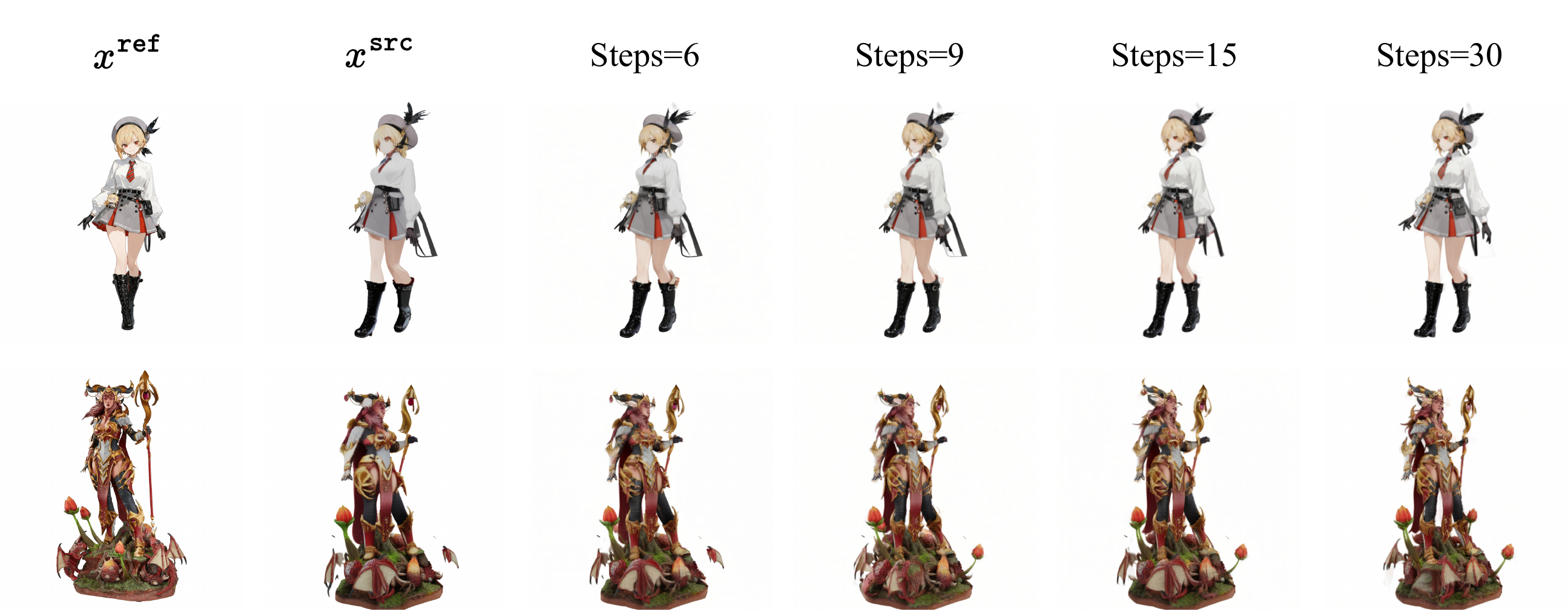}
    \caption{Effect of Editing Steps ($N_e$)}
  \end{subfigure}
  \captionof{figure}{Qualitative ablation study on RE parameters. (a) \textbf{Editing Ratio}: A low ratio (e.g., 0.25) limits editing capability, while a full ratio (1.0) destroys the original structure. Our choice of 0.75 strikes a balance. (b) \textbf{Editing Steps}: The examples show changes in appearance detail as the number of editing steps increases.}
  \label{fig:ablation_vis}
\end{figure*}
\textbf{Ablation Study on RE Parameters.}
We study two different RE parameters separately: the editing ratio $N_e/N$, which controls how much of the full reverse-time schedule is used for editing, and the absolute number of editing steps $N_e$, which controls how many updates are performed within that editing interval.
\textbf{Editing Ratio.} With the schedule length fixed, a ratio of $0.25$ applies editing only over a short interval and therefore produces limited detail enhancement, while a ratio of $1.0$ edits the entire trajectory and causes visible changes in pose, scale, and identity.
The intermediate ratio of $0.75$ improves texture fidelity while preserving the source silhouette and viewpoint, as reflected by the trends in Fig.~\ref{fig:ablation_ratio_steps}(a) and the examples in Fig.~\ref{fig:ablation_vis}(a).
\textbf{Editing Steps.} We then vary $N_e$ while keeping the editing ratio approximately fixed. More steps improve CLIP/DINO similarity but reduce Mask IoU and increase computation, with diminishing visual gains beyond $9$ steps, as shown in Fig.~\ref{fig:ablation_ratio_steps}(b) and Fig.~\ref{fig:ablation_vis}(b).
We therefore use $N_e=9$ and $N=12$ in our 3D experiments to balance improved visual fidelity, source-view structural consistency, and practical editing cost.

\setcounter{figure}{3}
\begin{figure}[!tbp]
  \centering
  \includegraphics[width=\linewidth]{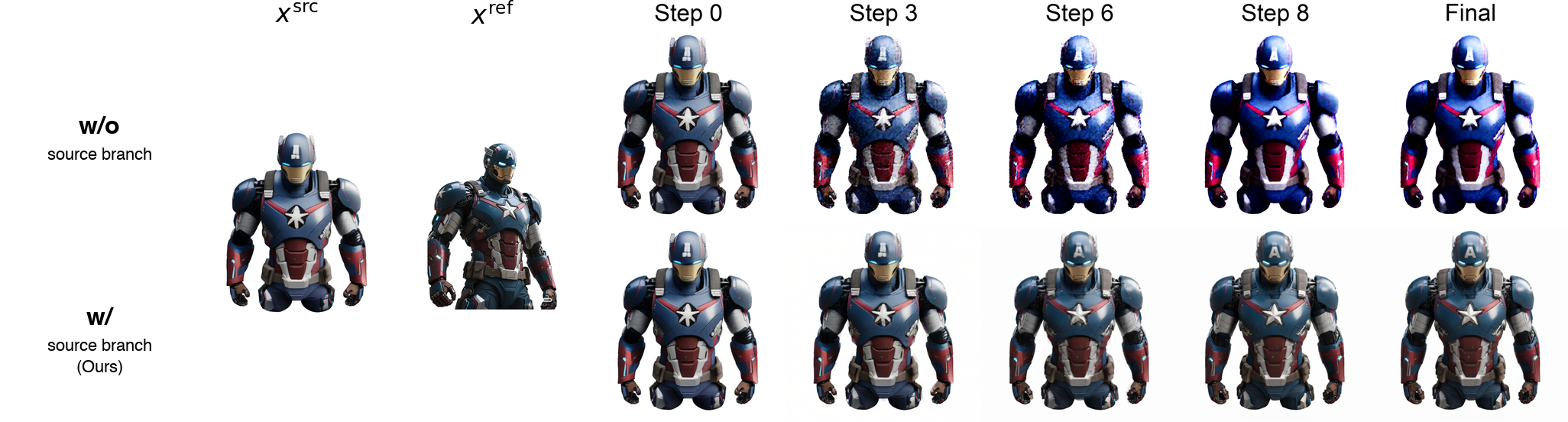}
  \caption{\textbf{Editing trajectories with and without the source branch.} Removing the source branch (top) leads to progressive color over-saturation and structural drift; keeping it (bottom, Ours) stabilizes the trajectory. Details in Sec.~\ref{sec:exp_feedback}.}
  \label{fig:neg_src_guidance}
\end{figure}

\begin{table}[!tbp]
  \centering
  \caption{\textbf{Ablation on key components of Reinforced Editing.} We report $\Delta$CLIP and $\Delta$DINO, defined as the change in CLIP/DINO similarity to $x^{\texttt{ref}}$ before and after editing; higher is better. Details are in Sec.~\ref{sec:exp_feedback}.}
  \label{tab:ablation_components}
  \begin{tabular}{@{}lcc@{}}
    \toprule
    Variant & $\Delta$CLIP Sim. $\uparrow$ & $\Delta$DINO Sim. $\uparrow$ \\
    \midrule
    Full (Ours) & \textbf{+0.0379} & \textbf{+0.0298} \\
    w/o Source Branch & $-0.0523$ & $-0.0497$ \\
    w/o Noise Update & $-0.0001$ & $+0.0089$ \\
    \bottomrule
  \end{tabular}
\end{table}

\textbf{Ablation Study on RE Components.}
RE consists of a source-aware regularization term and a dynamic noise update.
Without the source branch, target-conditioned guidance causes progressive color over-saturation and structural drift, as shown in Fig.~\ref{fig:neg_src_guidance}.
Without the noise update, the editing effect becomes much weaker and yields only marginal visual improvement.
Table~\ref{tab:ablation_components} reports the corresponding changes in CLIP/DINO similarity, confirming that both components contribute to the desired editing effect.
Together, these components provide effective and controlled feedback for downstream 3D supervision.

\setcounter{figure}{6}
\begin{figure*}[!t]
  \centering
  \includegraphics[width=\textwidth]{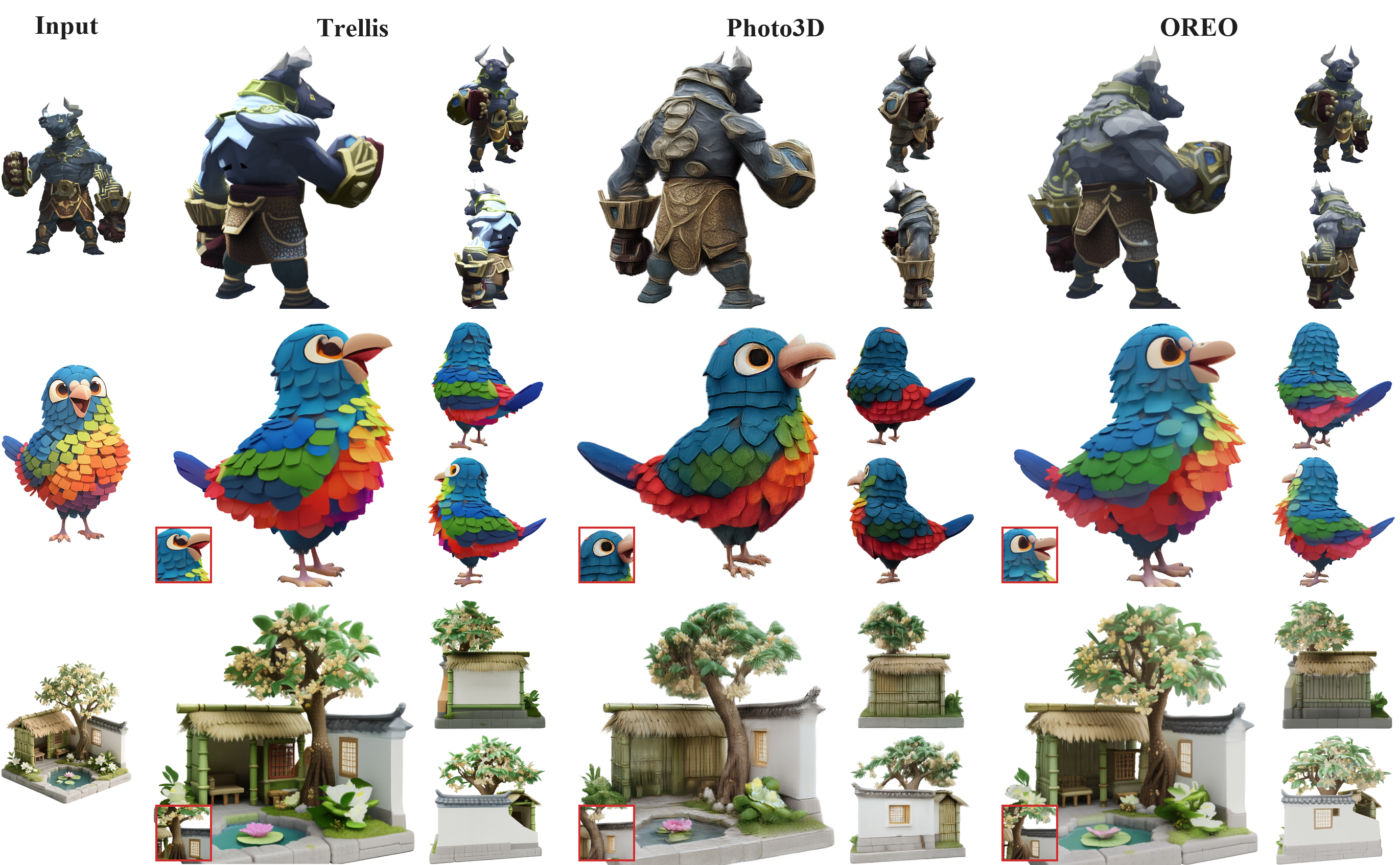}
  \caption{\textbf{Main qualitative comparison of 3D generation.} OREO delivers higher multi-view visual fidelity than Trellis and Photo3D. Details are in Sec.~\ref{sec:exp_main}.}
  \label{fig:qualitative_comparison}
  \vspace{2mm}
  \begin{minipage}{0.68\textwidth}
    \centering
    \captionof{table}{\textbf{Main quantitative comparison of 3D generation.} CLIP and DINO similarity to $x^{\texttt{ref}}$ on the Conceptual Design Dataset and GSO; see Sec.~\ref{sec:exp_main}.}
    \label{tab:main_results}
    \resizebox{\linewidth}{!}{%
      \begin{tabular}{@{}lcccc@{}}
        \toprule
        \multirow{2}{*}{Method} & \multicolumn{2}{c}{Conceptual Design} & \multicolumn{2}{c}{GSO} \\
        \cmidrule(lr){2-3}\cmidrule(lr){4-5}
         & CLIP Sim. $\uparrow$ & DINO Sim. $\uparrow$ & CLIP Sim. $\uparrow$ & DINO Sim. $\uparrow$ \\
        \midrule
        Trellis & 0.7613 & 0.7916 & 0.7722 & 0.7022 \\
        Photo3D & 0.7380 & 0.7837 & 0.7512 & 0.7034 \\
        OREO (Ours) & \textbf{0.7834} & \textbf{0.8065} & \textbf{0.7764} & \textbf{0.7069} \\
        \bottomrule
      \end{tabular}}
  \end{minipage}
\end{figure*}
\subsection{Main Results}
\label{sec:exp_main}

\textbf{Experimental Setup.}
We evaluate the 3D generation quality of Trellis distilled with OREO. 
As baselines, we compare against the pre-trained Trellis and Photo3D~\cite{liang2026photo3d}, an offline detail-enhancement method built on the same Trellis backbone.
Following the protocol in Sec.~\ref{sec:exp_feedback}, we render five views for each reference at azimuths $-90^\circ$, $-45^\circ$, $0^\circ$, $45^\circ$, and $90^\circ$.
We report CLIP-ViT-L/14~\cite{radford2021learning} and DINOv3-ViT-L/16~\cite{simeoni2025dinov3} embedding similarity to the reference $x^{\texttt{ref}}$, averaged over the five views.
In addition to our proposed Conceptual Design Dataset, which targets imaginative and out-of-distribution references, we further validate on GSO~\cite{downs2022google}, an in-distribution benchmark of common real-world objects, to assess whether OREO maintains the pre-trained model's performance on its original distribution. This provides a complementary check of generalization.

\begin{figure*}[p]
  \centering
  \includegraphics[width=\textwidth]{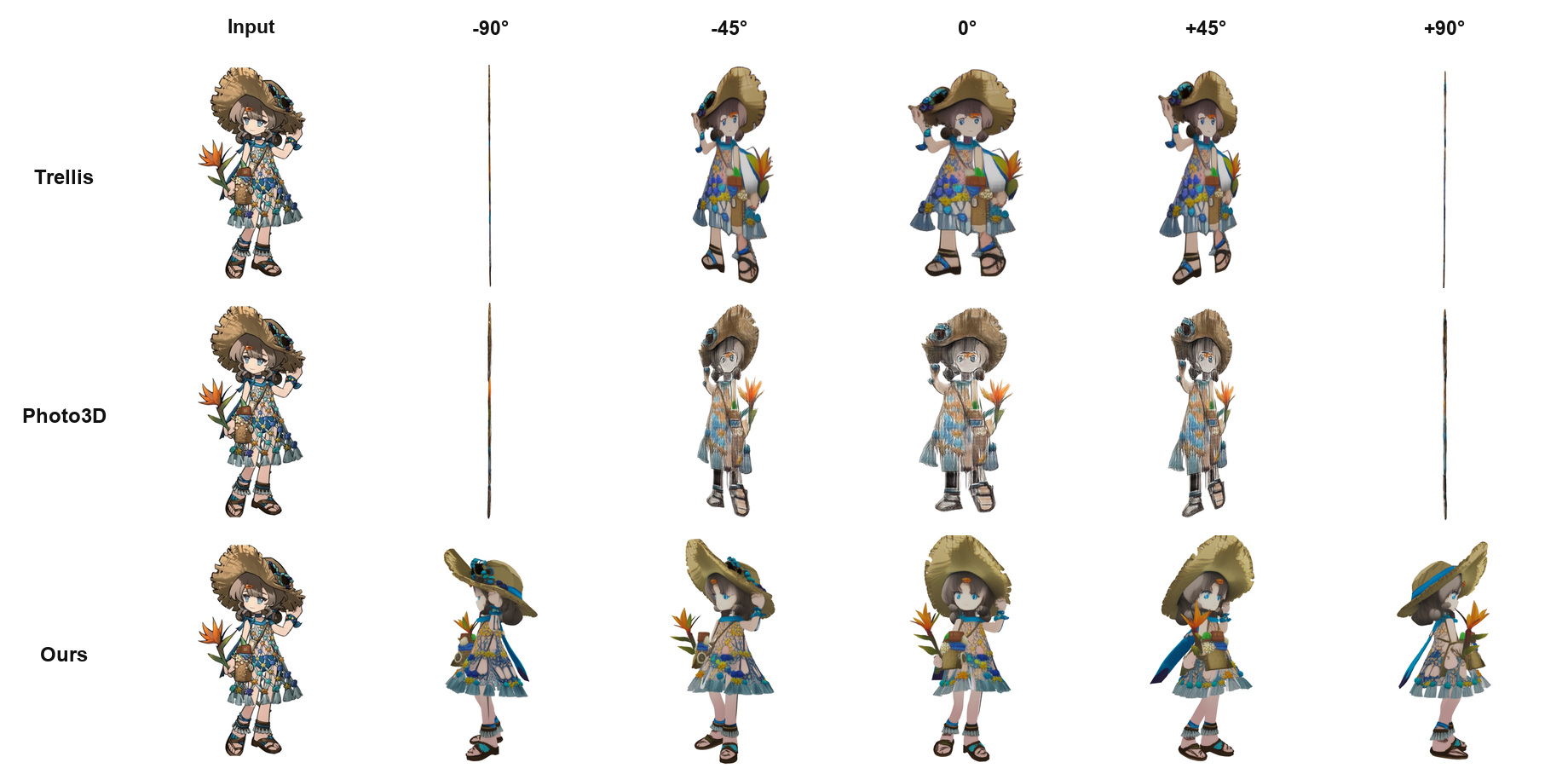}\par\vspace{-2mm}
  \includegraphics[width=\textwidth]{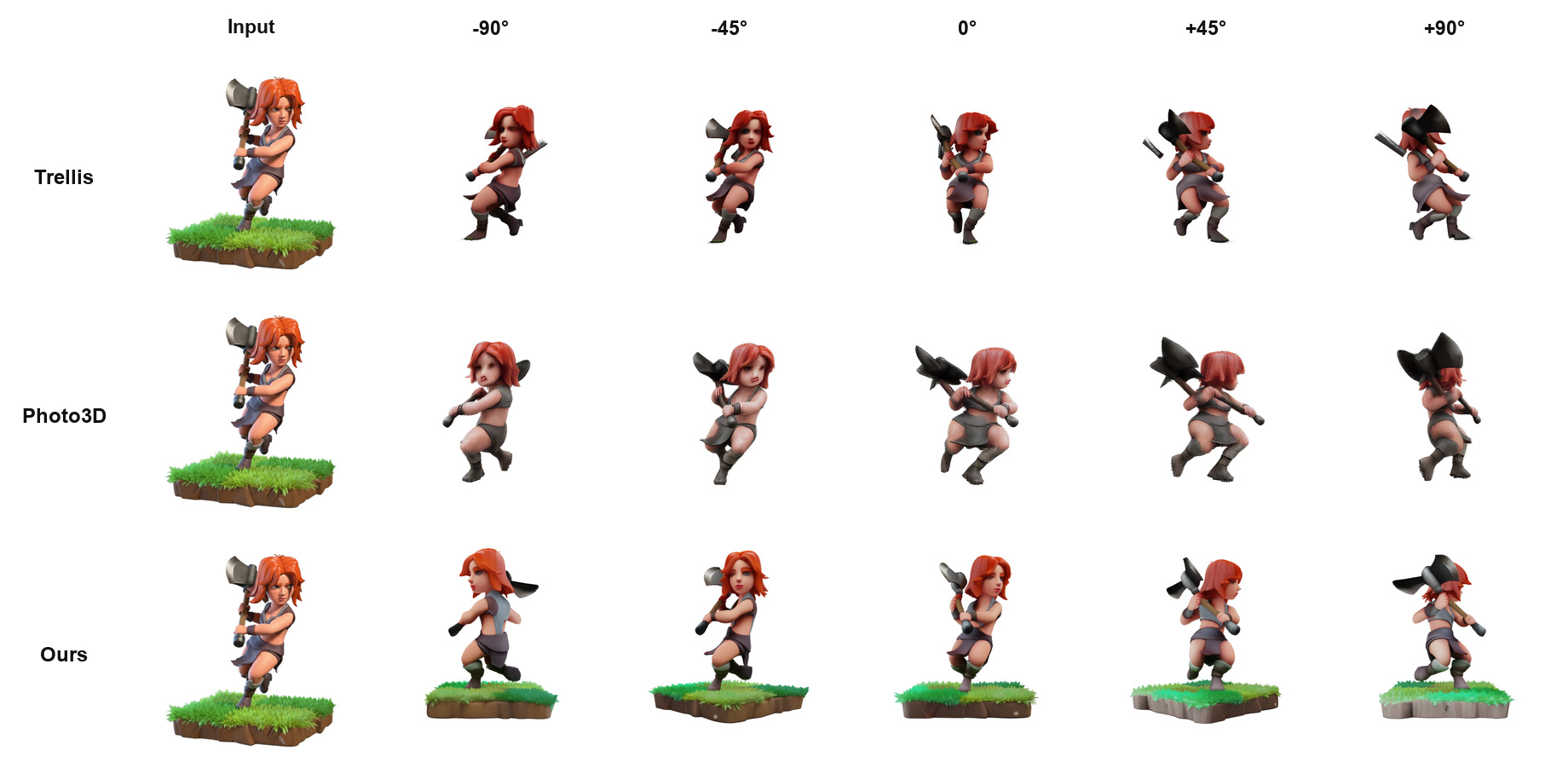}\par\vspace{-2mm}
  \includegraphics[width=\textwidth]{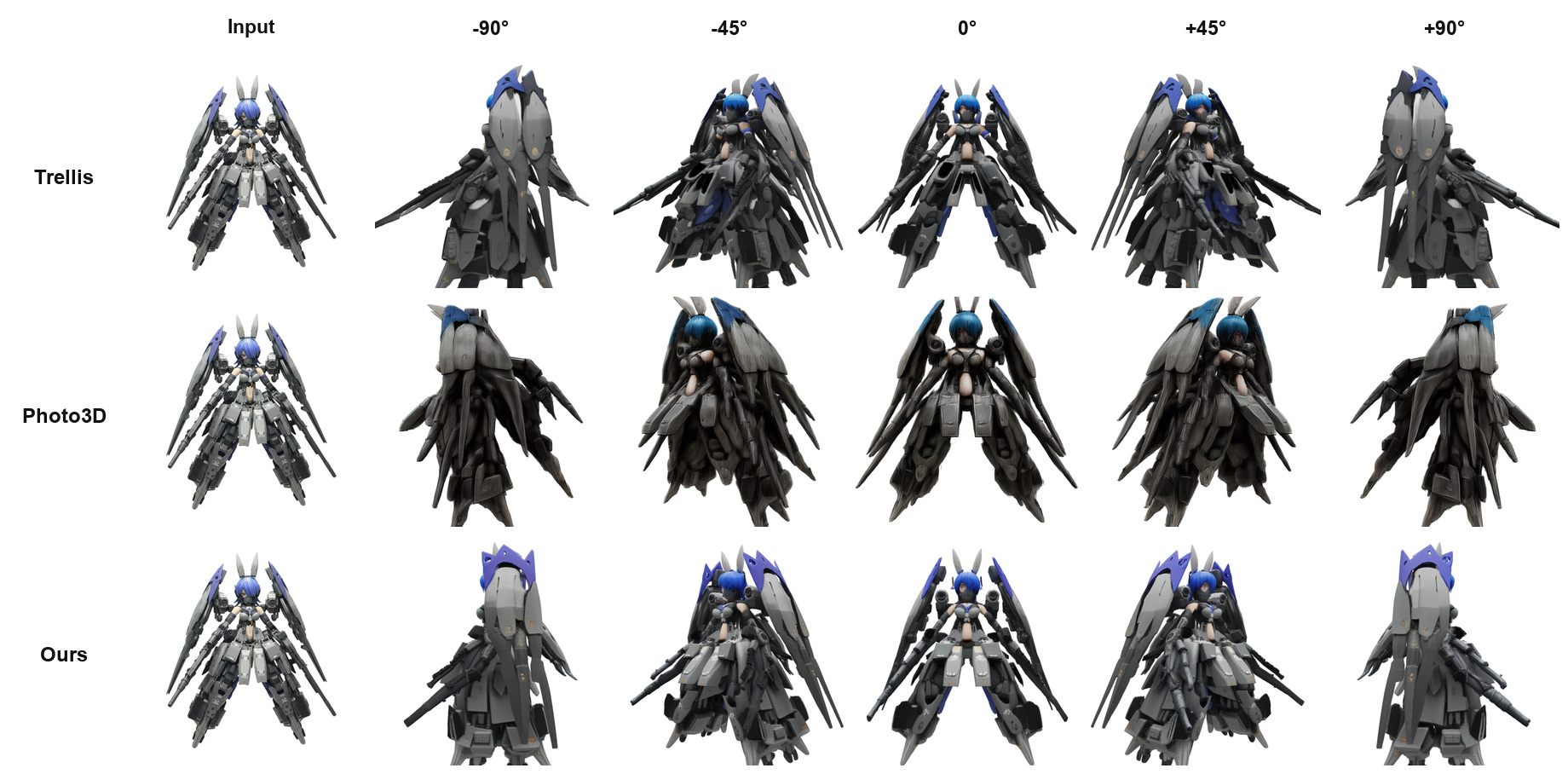}
  \caption{\textbf{Additional multi-view comparisons.} Each example compares Trellis, Photo3D, and OREO using the input reference and five views at azimuths $-90^\circ$, $-45^\circ$, $0^\circ$, $45^\circ$, and $90^\circ$.}
  \label{fig:additional_multiview_main}
\end{figure*}

\begin{figure*}[!t]
  \centering
  \includegraphics[width=0.88\textwidth]{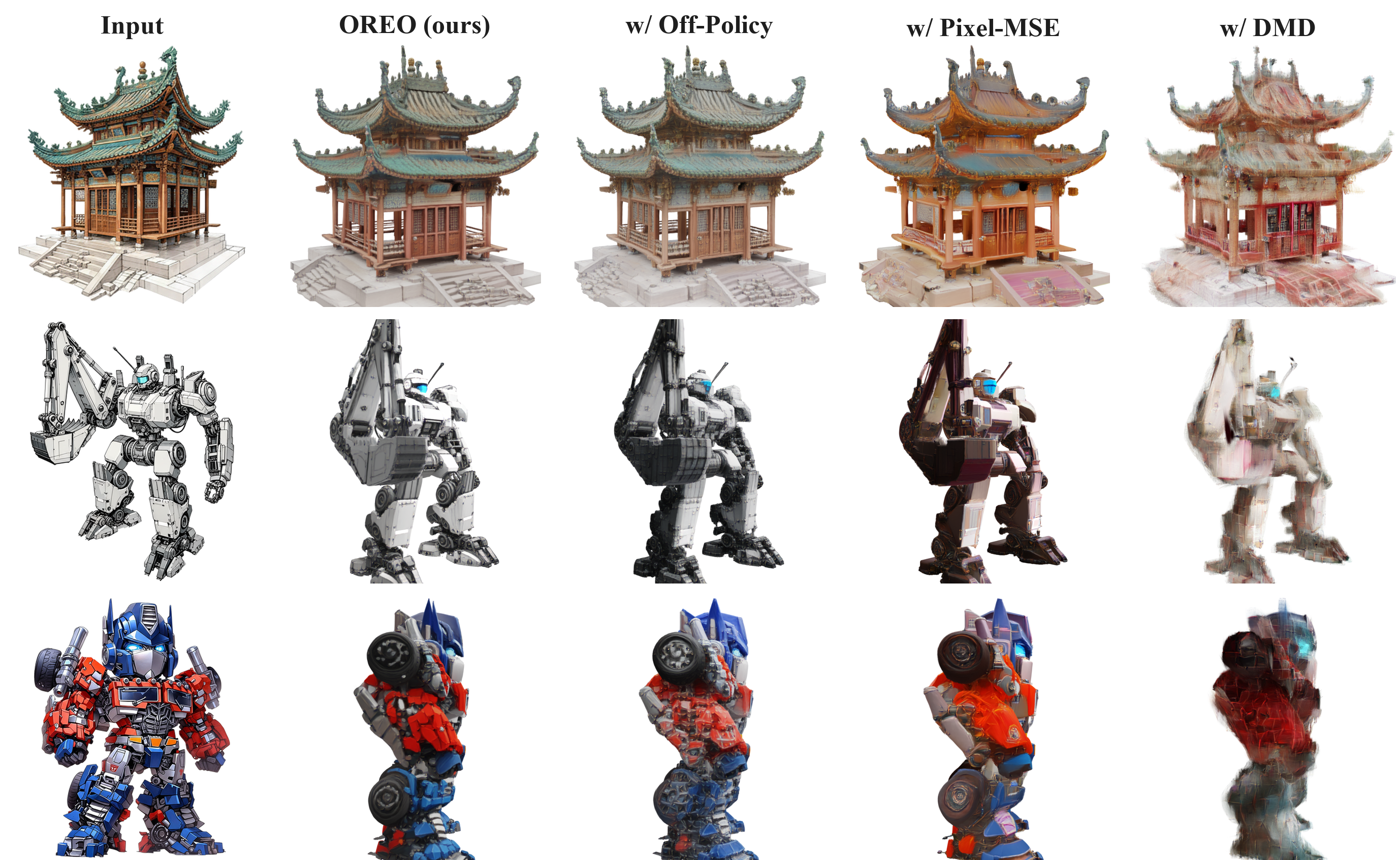}
  \caption{\textbf{Qualitative ablation of the 3D generator design choices.} We compare Full OREO against other training variants. Details are in Sec.~\ref{sec:exp_ablation}.}
  \label{fig:ablation_overview}
  \vspace{2mm}
  \begin{minipage}{0.68\textwidth}
    \centering
    \captionof{table}{\textbf{Ablation on the 3D generator design choices.} CLIP and DINO similarity to $x^{\texttt{ref}}$ on the Conceptual Design Dataset; see Sec.~\ref{sec:exp_ablation}.}
    \label{tab:ablation_quantitative}
    \begin{tabular}{@{}lcc@{}}
      \toprule
      Method / Variant & CLIP Sim. $\uparrow$ & DINO Sim. $\uparrow$ \\
      \midrule
      Pretrained Trellis & 0.7613 & 0.7916 \\
      \midrule
      \textbf{Full OREO (Ours)} & \textbf{0.7834} & \textbf{0.8065} \\
      w/ Pixel-MSE Supervision & 0.6617 & 0.6982 \\
      w/ Off-policy Rollout & 0.7279 & 0.7613 \\
      w/ DMD (Score Distillation) & 0.5393 & 0.5486 \\
      \bottomrule
    \end{tabular}
  \end{minipage}
\end{figure*}
\textbf{Quantitative and Qualitative Evaluation.}
Table~\ref{tab:main_results} summarizes the quantitative comparison. OREO achieves clear gains in CLIP and DINO similarity on the Conceptual Design Dataset, our target imaginative distribution, while maintaining comparable performance on GSO.
Fig.~\ref{fig:qualitative_comparison} shows the visual comparison. Compared with the pre-trained Trellis and Photo3D, OREO recovers finer details while preserving geometry and identity across views.
Fig.~\ref{fig:additional_multiview_main} provides more qualitative comparisons across views. Unlike Photo3D's detail-focused post-training, OREO updates Trellis's sparse-structure stage to enable such shape corrections. More examples are provided in \suppref{sec:additional_multiview}.
These results show that the improvements produced by Reinforced Editing can be effectively distilled into the 3D generator. Although each iteration provides only a limited update, these improvements accumulate over training and lead to substantial gains in both appearance and shape. The distilled supervision is applied to both stages of the Trellis generator, including sparse structure generation and sparse feature generation, allowing OREO to progressively improve the shape and appearance of generated assets across multiple viewpoints.

\textbf{User Preference Study.}
To complement the automatic metrics, we further conduct a user preference study on the same held-out Conceptual Design Dataset, in which participants compare multi-view renderings from Pretrained Trellis, Photo3D, and OREO under blinded conditions.
OREO receives the highest aggregate preference share, with 38\% of all responses.
The \suppref{sec:user_study_details} provides the full study protocols and preference results.


\subsection{Ablation Study and Analysis}
\label{sec:exp_ablation}

Full OREO combines on-policy rollouts, latent contrastive supervision, and explicit RE-edited renderings as 2D pseudo-targets. We compare it with off-policy rollout, pixel MSE, and DMD~\cite{yin2024onestep} supervision. Table~\ref{tab:ablation_quantitative} shows that all three variants underperform pretrained Trellis, while Full OREO improves both CLIP and DINO similarity on the same held-out references in this comparison.

\textbf{Degradation under Off-policy Rollout.} Fixing the rollout to the pretrained generator's output ties supervision to its initial distribution rather than the generator's current trajectory. As the student evolves, this fixed supervision becomes stale and cannot correct newly visited states. Fig.~\ref{fig:ablation_overview} shows that the geometry remains largely preserved, but colors desaturate and fine-grained textures are smoothed out. Together with the lower scores in Table~\ref{tab:ablation_quantitative}, this supports on-policy rollouts as a way to keep supervision aligned with the evolving generator throughout training.

\textbf{Attenuation of Gradient through Differentiable Rendering.} This variant keeps the RE-edited views but replaces latent contrastive supervision with pixel-space regression $\|x^{\texttt{src}}-x^{\texttt{tgt}}\|^2$, sending the loss through the decoder $\mathcal{D}$ and renderer $\mathcal{P}$. It also loses the positive/negative latent comparison between fidelity correction and preserved source content. Pixel MSE therefore treats all image discrepancies as direct regression errors, without distinguishing details to enhance from content to preserve. The outputs in Fig.~\ref{fig:ablation_overview} become texture-averaged and color-shifted, and Table~\ref{tab:ablation_quantitative} shows lower CLIP and DINO scores. This comparison supports latent contrastive supervision as a more effective way to transfer the RE correction to the 3D generator.

\textbf{Role of Reinforced Editing.} The DMD variant replaces the explicit RE-edited rendering target with an image-editor DMD loss, giving an implicit score signal through the renderer and decoder. Unlike RE, this signal provides no concrete, structure-preserving target specifying appearance changes while keeping viewpoint and source content fixed. Fig.~\ref{fig:ablation_overview} shows geometric distortion and color collapse, while Table~\ref{tab:ablation_quantitative} reports the lowest scores. This supports explicit RE-edited renderings as stable supervision for the downstream 3D generator.

\section{Conclusion}
This paper proposes \textbf{OREO}, a fidelity alignment framework for 3D generation.
Addressing the lack of visual realism in existing 3D models, we introduce an on-the-fly optimization loop that leverages 2D diffusion priors.
In the Render--Edit--Optimize loop, Reinforced Editing produces structure-preserving 2D pseudo-targets. Contrastive Distillation uses edited and source renderings as conditions to construct positive and negative latent targets for direct generator supervision.
This design provides dynamic supervision without paired 3D ground truth, refreshing targets from the evolving generator while retaining the pretrained 3D prior for geometric stability.
Experimental results show that OREO improves the visual fidelity and texture detail of 3D generators while largely preserving structure.
Limitations are provided in \suppref{sec:limitations_failure}.

\clearpage

\bibliographystyle{splncs04}
\bibliography{main}

\clearpage
\def\thefootnote{\arabic{footnote}}
\setcounter{footnote}{0}
\setcounter{section}{0}
\setcounter{figure}{0}
\setcounter{table}{0}
\setcounter{equation}{0}
\setcounter{algorithm}{0}
\renewcommand{\thesection}{S\arabic{section}}
\renewcommand{\thefigure}{S\arabic{figure}}
\renewcommand{\thetable}{S\arabic{table}}
\renewcommand{\theequation}{S\arabic{equation}}
\renewcommand{\thealgorithm}{S\arabic{algorithm}}
\renewcommand{\bibliographystyle}[1]{}
\renewcommand{\bibliography}[1]{}

\begin{center}
  {\Large\bfseries OREO: Fidelity Alignment in 3D Generation via\par}
  {\Large\bfseries On-the-fly Rendering-Editing Optimization\par}
  \vspace{1mm}
  {\large\bfseries Supplementary Material\par}
\end{center}
\markboth{}{OREO: Supplementary Material}
\thispagestyle{empty}
\vspace{-3mm}

\section*{Overview}
This supplementary material first presents additional multi-view comparisons among OREO, the pretrained Trellis, and Photo3D. We then provide details on dataset construction, training and editing configurations, the user preference study protocol, computational overhead, and limitations and failure cases.

\section{Additional Multi-view Comparisons}
\label{sec:additional_multiview}
\begin{figure}[H]
  \centering
  \includegraphics[width=0.82\textwidth]{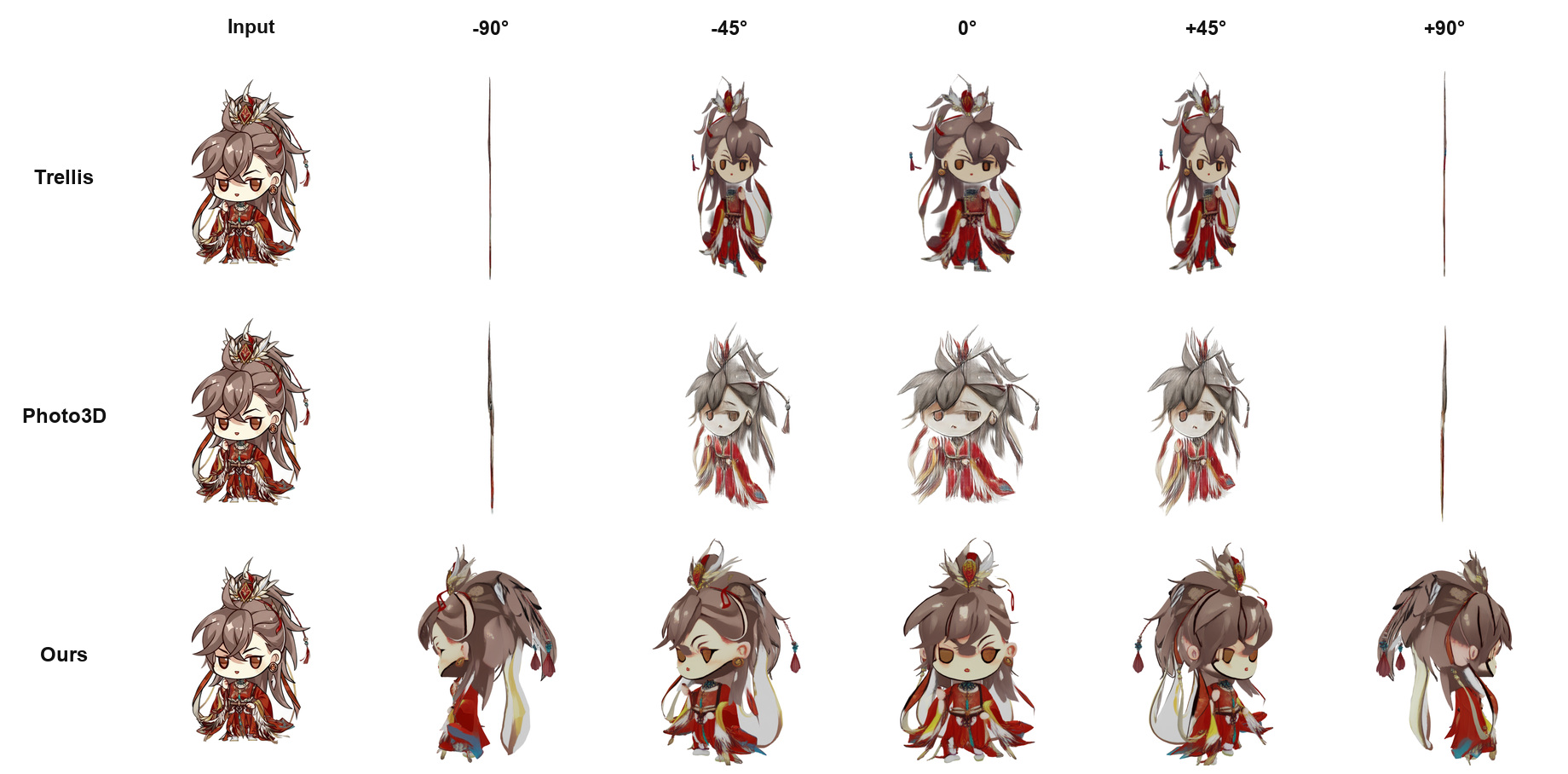}
  \vspace{-2mm}
  \includegraphics[width=0.82\textwidth]{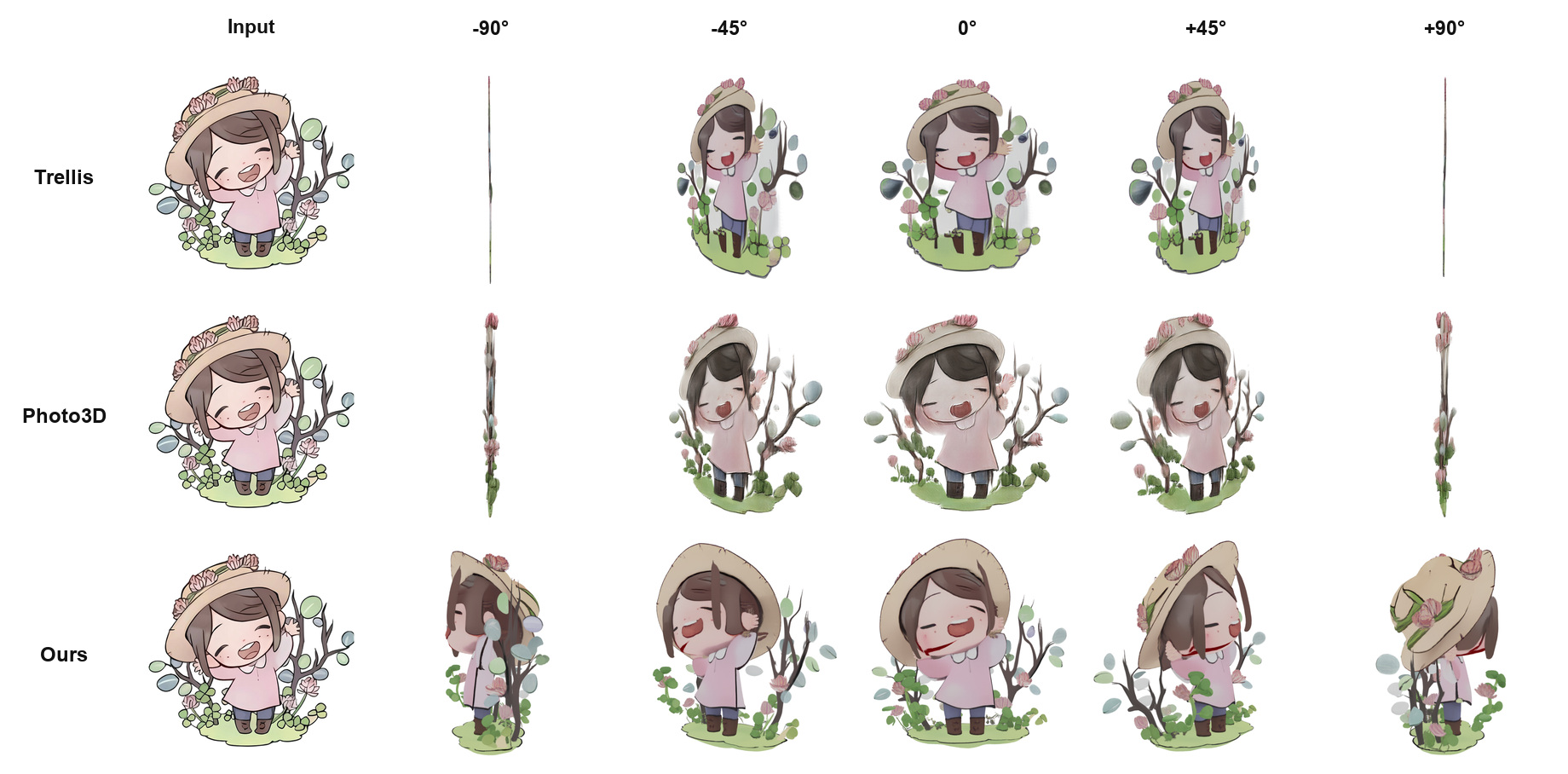}
  \caption{Additional multi-view comparisons of the pretrained Trellis, Photo3D, and OREO. Each block shows the input reference, followed by five views rendered at the azimuth angles $-90^\circ$, $-45^\circ$, $0^\circ$, $45^\circ$, and $90^\circ$.}
  \label{fig:additional_multiview_1}
\end{figure}

\clearpage
\begin{figure}[H]
  \centering
  \includegraphics[width=\textwidth]{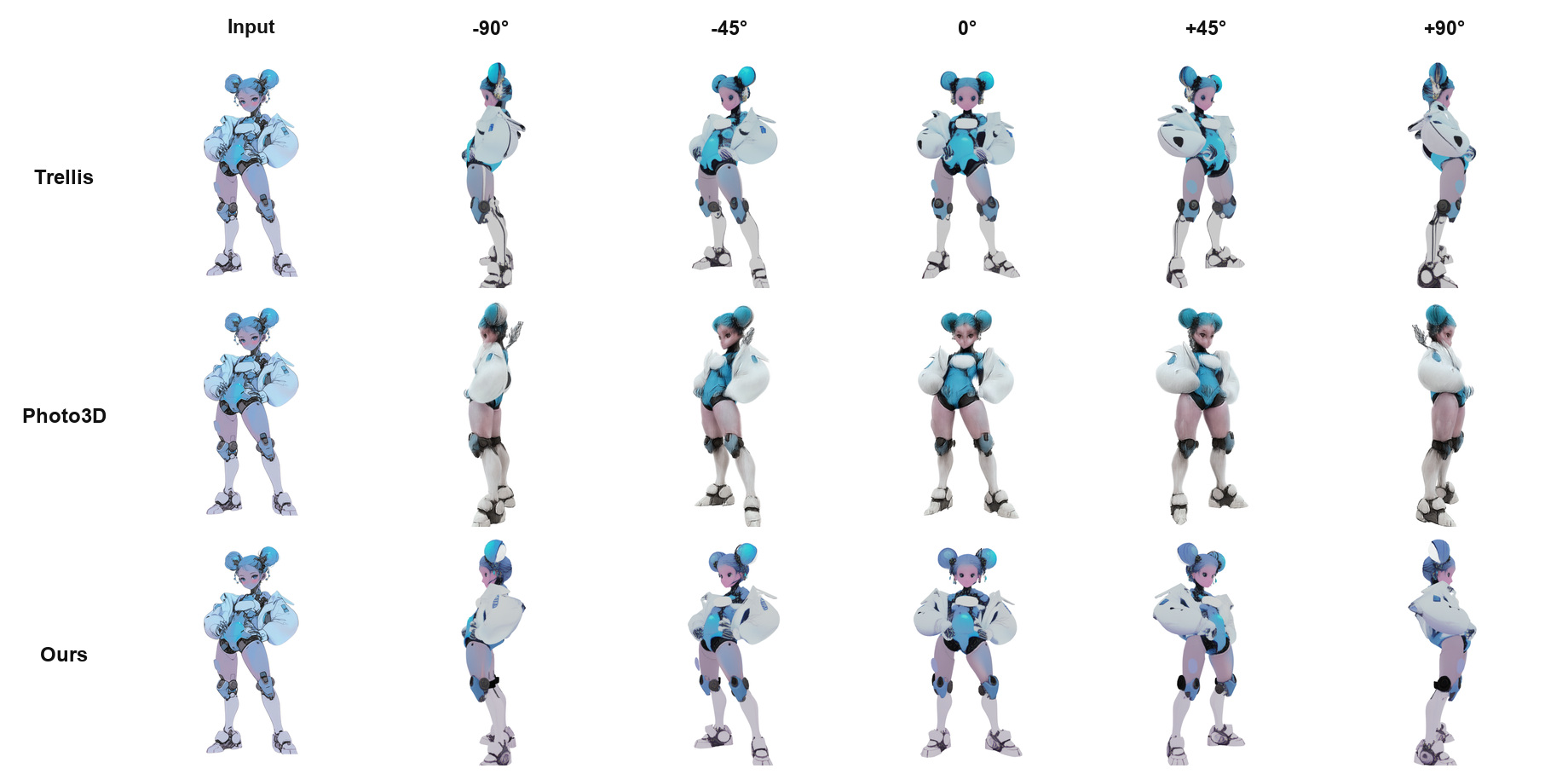}
  \vspace{-2mm}
  \includegraphics[width=\textwidth]{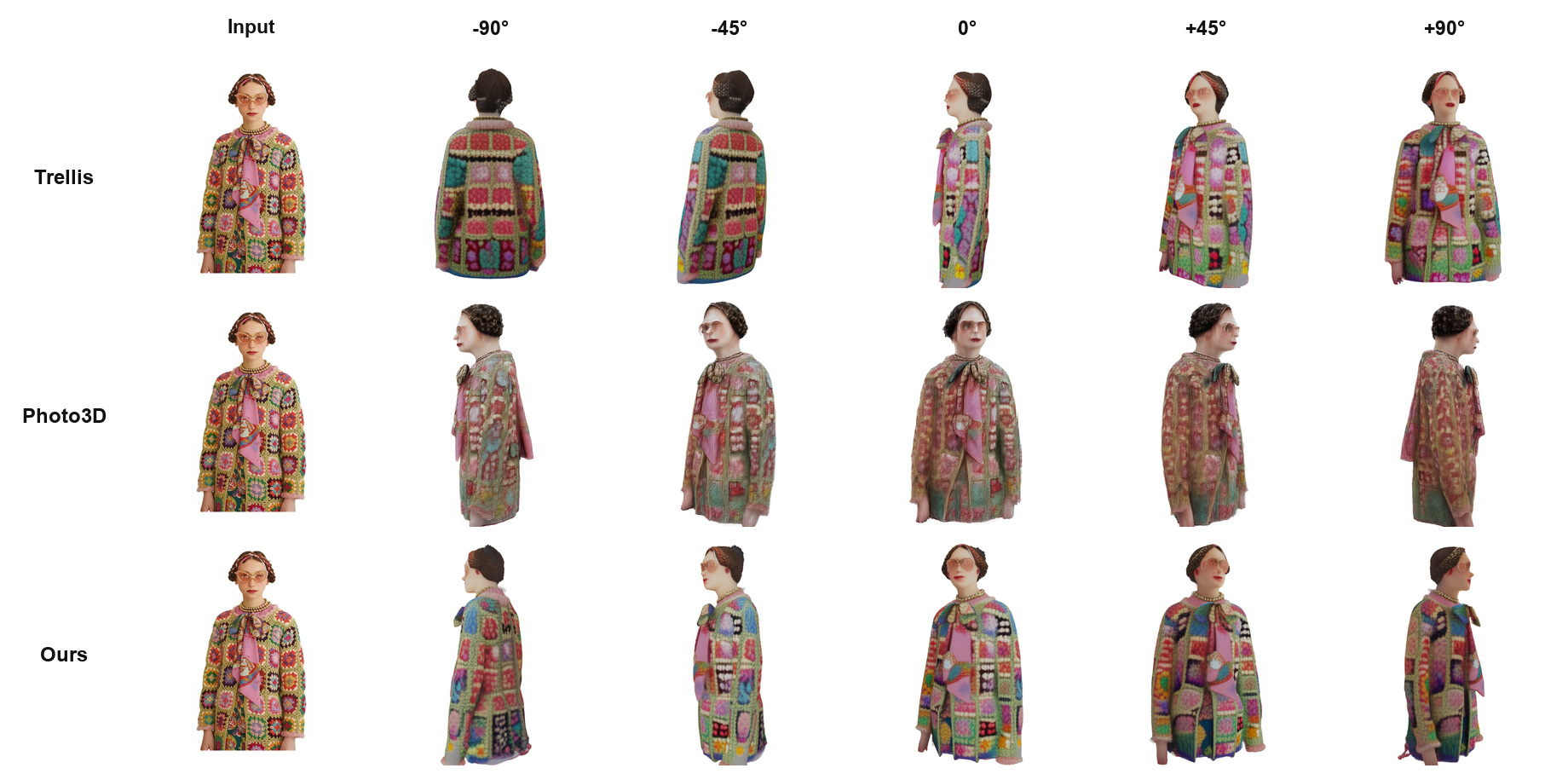}
  \vspace{-2mm}
  \includegraphics[width=\textwidth]{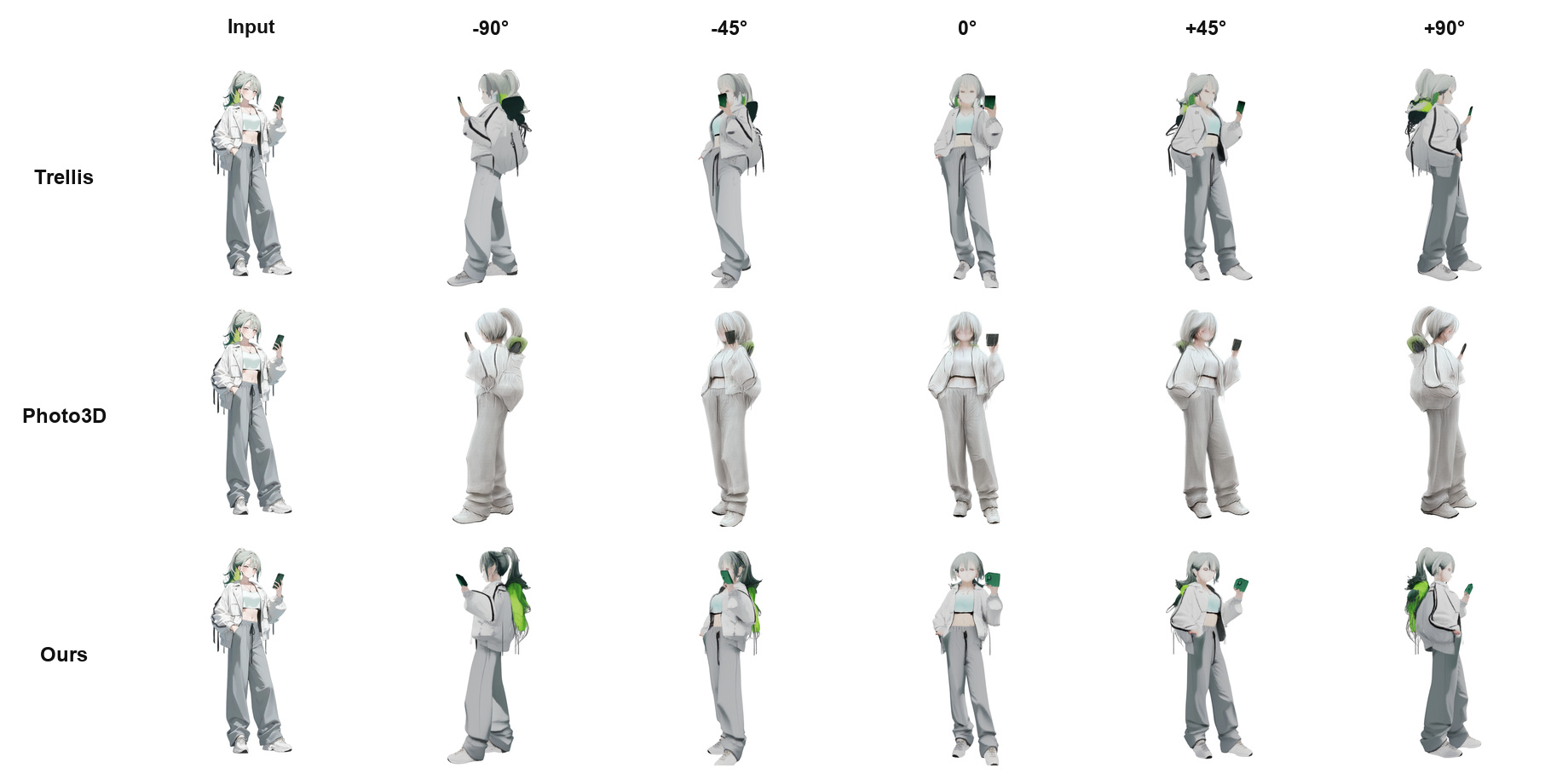}
  \caption{Additional multi-view qualitative comparisons (continued).}
  \label{fig:additional_multiview_2}
\end{figure}

\clearpage
\begin{figure}[H]
  \centering
  \includegraphics[width=\textwidth]{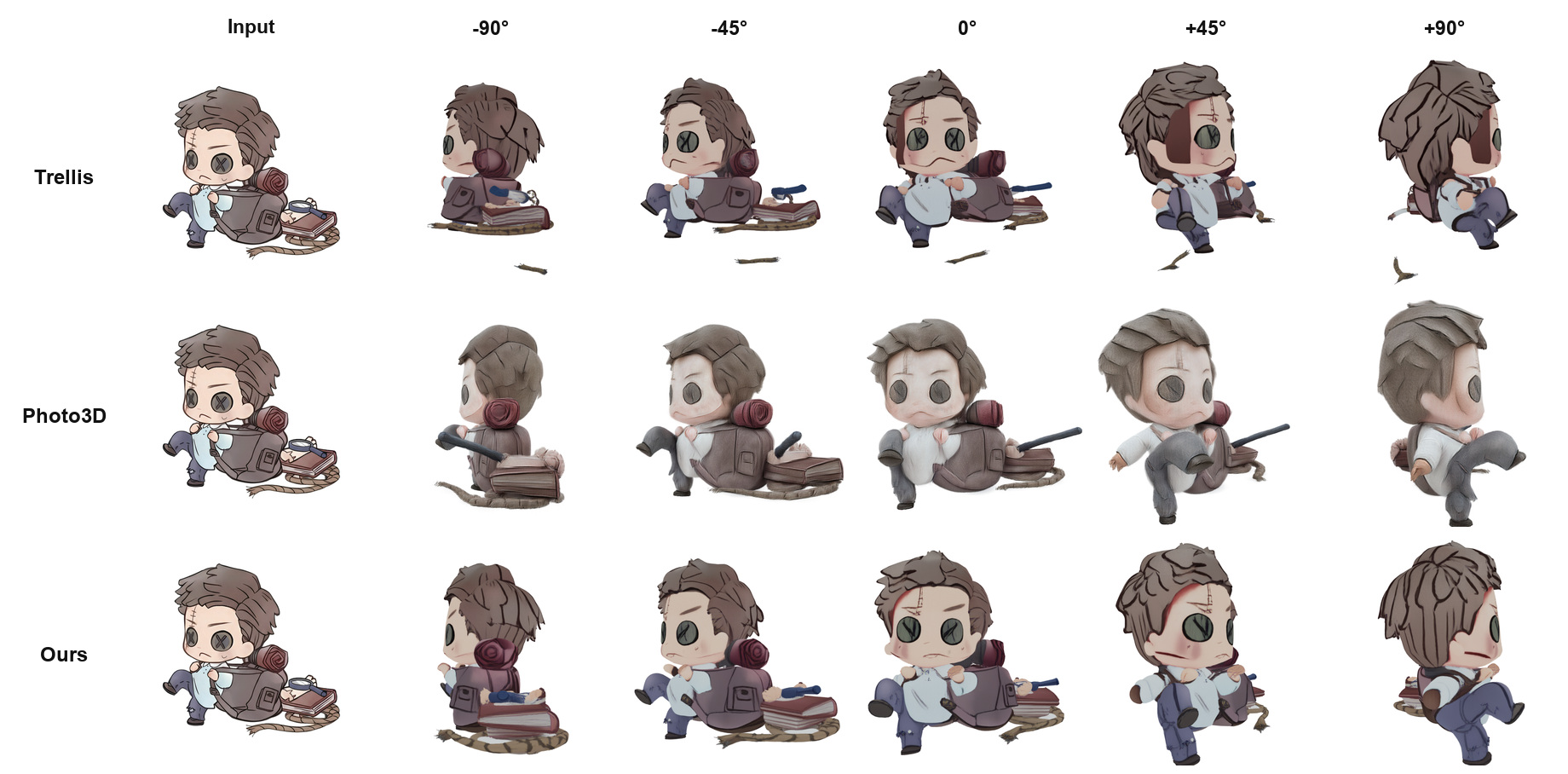}
  \vspace{-2mm}
  \includegraphics[width=\textwidth]{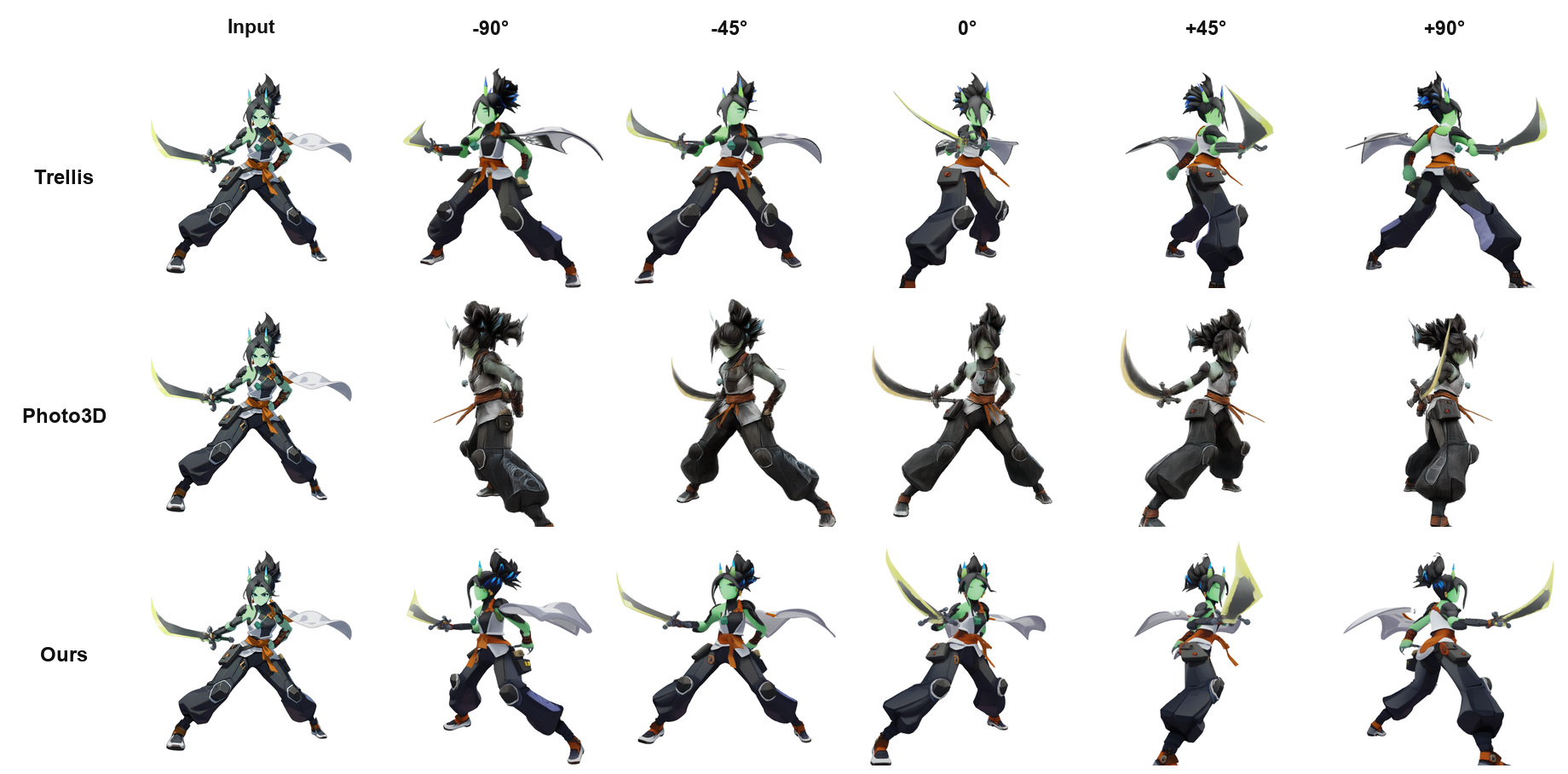}
  \vspace{-2mm}
  \includegraphics[width=\textwidth]{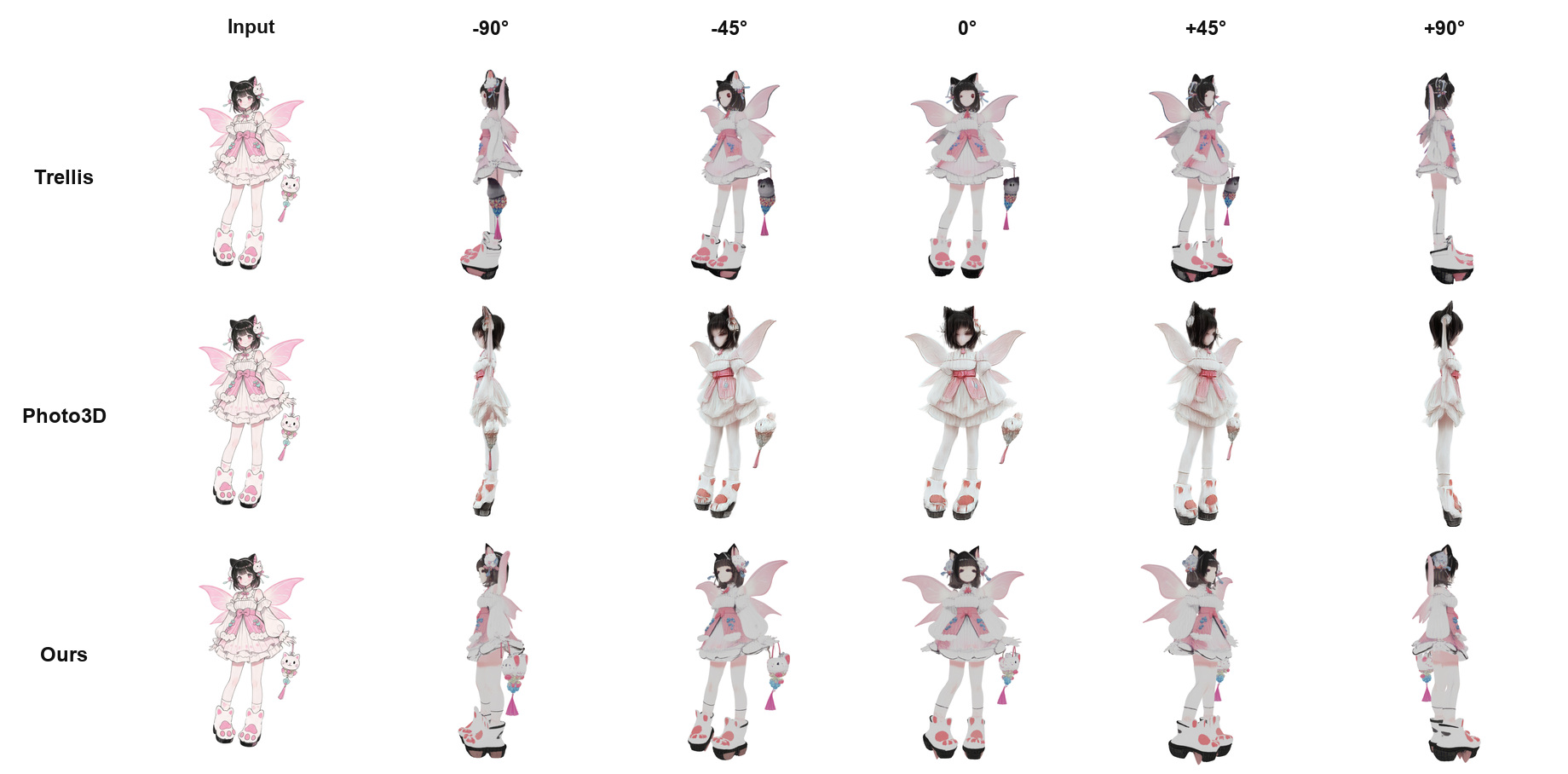}
  \caption{Additional multi-view qualitative comparisons (continued).}
  \label{fig:additional_multiview_3}
\end{figure}

\clearpage
\begin{figure}[H]
  \centering
  \includegraphics[width=\textwidth]{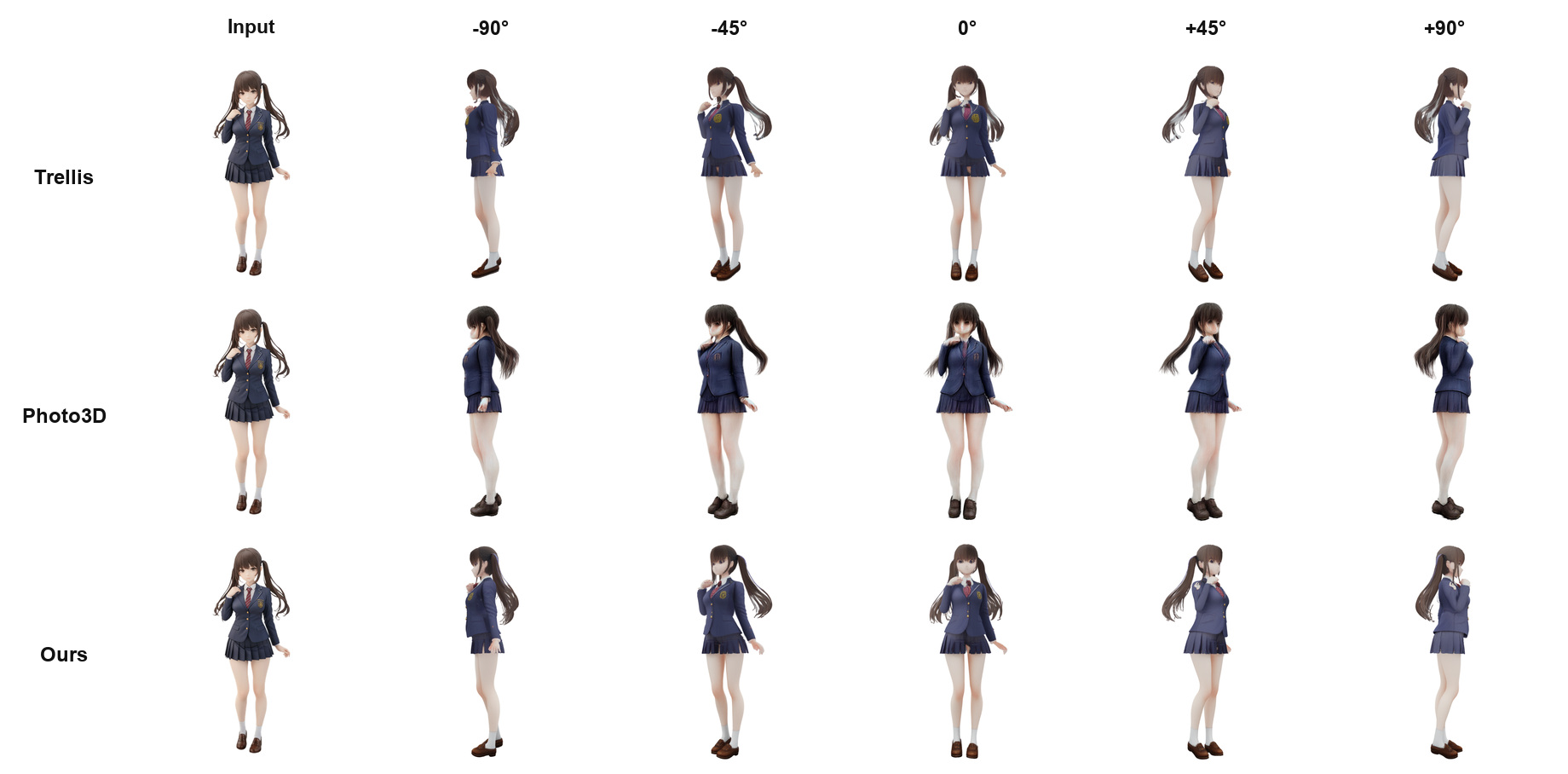}
  \vspace{-2mm}
  \includegraphics[width=\textwidth]{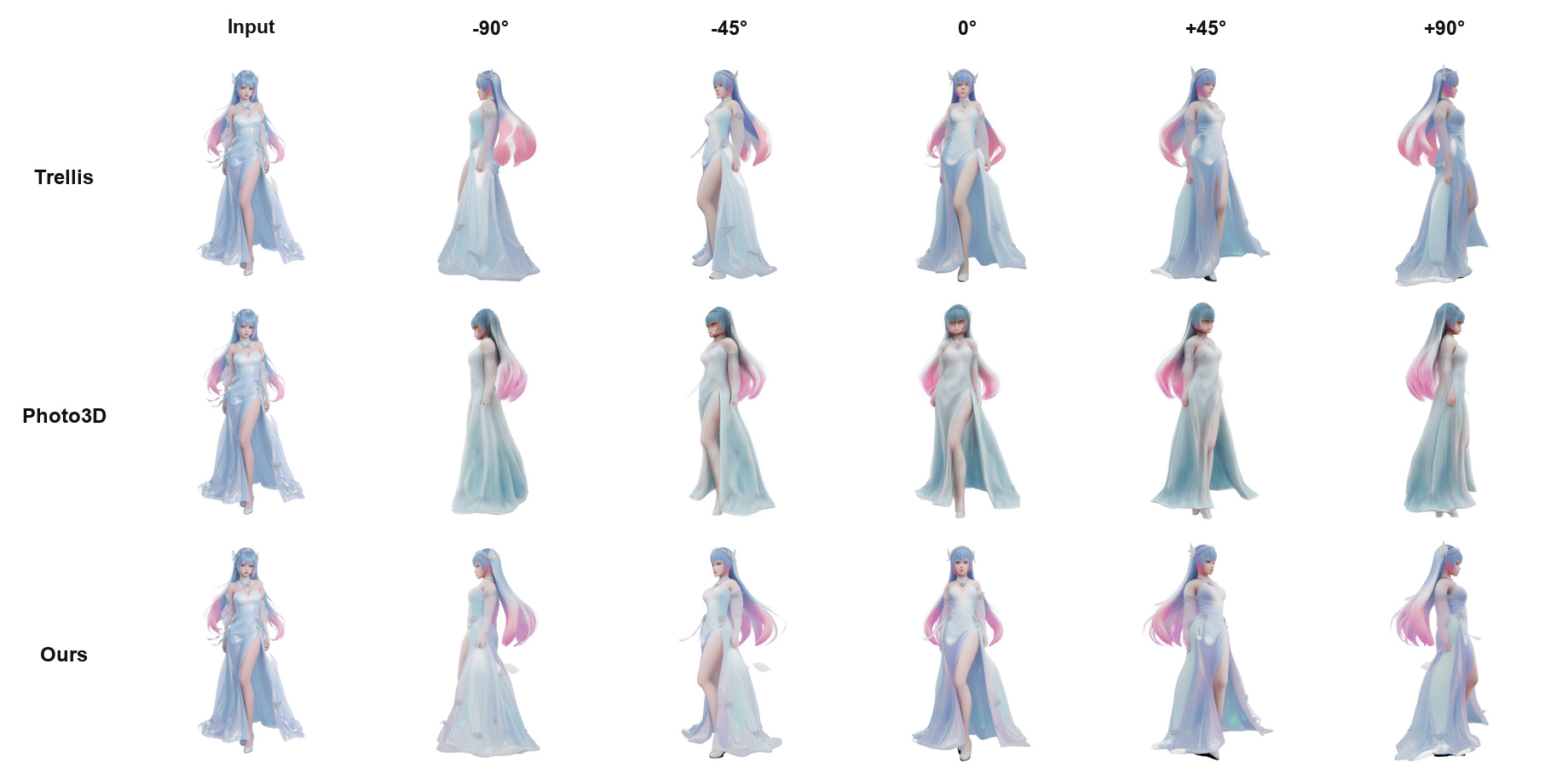}
  \caption{Additional multi-view qualitative comparisons (continued).}
  \label{fig:additional_multiview_4}
\end{figure}

\section{Implementation and Dataset Details}
\label{sec:implementation_details}
This section details dataset construction, training hyper-parameters, and the editing setup for the methods compared in our main experiments.

\subsection{Conceptual Design Dataset Construction}
\label{sec:dataset_construction}
We construct an image-only collection to support generator post-training without paired 3D supervision.
The dataset comprises 2,396 concept-design reference images that we manually curate from public online sources; most images are AI-generated or user-shared creations posted on public content platforms, reflecting the imaginative and stylized inputs that image-to-3D users bring in practice.
We prioritize references with clean foregrounds, well-defined silhouettes, and informative texture or material cues, and manually filter the collection to remove duplicates, low-resolution thumbnails, watermarked images, and content with unclear foregrounds.
We will release the dataset on our project page.

\subsection{Training Details}
\label{sec:training_details}
We post-train Trellis-image-large with OREO on the Conceptual Design Dataset training split, using the frozen Qwen-Image-Edit-2511~\cite{wu2025qwen} as our 2D editor.

\textbf{Optimization.}
We use the Adan optimizer with learning rate $10^{-4}$, weight decay $0$, and $\epsilon=10^{-4}$.
We use a per-GPU batch size of $1$ with no gradient accumulation, giving an effective batch size of $8$ reference images per iteration.
Full-parameter updates are applied to both Trellis generation stages, while the decoder $\mathcal{D}$ remains frozen.
Stage 1 generates the sparse structure from a latent represented on a dense 3D grid, and Stage 2 generates sparse SLAT features at the active coordinates.
We jointly post-train both flow models in all experiments.

\textbf{Rollout and camera sampling.}
Each iteration performs an on-policy ODE rollout with the current student generator to obtain a clean latent $z_0 = G_\theta(x^{\texttt{ref}})$.
The ODE rollout follows the sampling configuration of the pretrained Trellis model.
For every reference we sample a single training view with yaw uniformly drawn from $[0^\circ, 360^\circ]$, pitch fixed at $0^\circ$, and field of view $40^\circ$.
The camera distance is determined by an adaptive-distance rule with a fill ratio of $0.9$, so that the object occupies a consistent portion of the rendered view.
Rendering uses the Gaussian-splatting renderer at $1024\times1024$ resolution with a white background.

\textbf{Loss and time-step sampling.}
For the contrastive objective in Eq.~11 of the main paper, the training time step is sampled from $t \sim \mathcal{U}(0.02, 0.98)$.
The student prediction and the frozen pretrained-generator passes used to construct $z_0^{+}$ and $z_0^{-}$ all use classifier-free guidance with the default settings of the pretrained Trellis model.
The positive and negative passes share the same re-noised latent $\hat{z}_t$ and fresh noise draw $\epsilon'$, while using $x^{\texttt{tgt}}$ and $x^{\texttt{src}}$ as their respective visual conditions.
The total loss is $\mathcal{L}_{total} = \mathcal{L}_{contrast} + \lambda\,\mathcal{L}_{reg}$, with $\lambda = 1.0$.
The velocity regularizer $\mathcal{L}_{reg}$ compares the student and pretrained velocity predictions with the same noisy latent $z_t$, time step $t$, and reference condition $x^{\texttt{ref}}$.

\subsection{Editing Setup for Compared Methods}
\label{sec:editing_setup}
Reinforced Editing and the off-the-shelf 2D editors used as baselines share the same underlying editor backbone but differ in their text-conditioning interface.
Both prompt families are fixed across the dataset without per-example tuning.

\textbf{Reinforced Editing.}
We instantiate Reinforced Editing on top of Qwen-Image-Edit-2511 with $N=12$ total time steps and $N_e=9$ editing steps applied at the last portion of the schedule (editing ratio $0.75$).
The target branch uses classifier-free guidance with scale $w=4$, while the source branch is an unconditional forward pass, in accordance with Eq.~7 of the main paper.
The shared noise $\epsilon$ is initialized once per view and updated at each step by injecting the CFG signal, as in Eq.~8 of the main paper.
The editing resolution matches the renderer at $1024\times1024$.
The text component of $c^{\texttt{ref}}$ is:
\begin{quote}
\itshape
``Rotate the camera. White background.''
\end{quote}

\textbf{Prompt selection.}
We select a single fixed prompt by evaluating representative instructions on the Conceptual Design Dataset while keeping the editor and all RE hyper-parameters unchanged.
Table~\ref{tab:prompt_ablation} reports the change in CLIP and DINO similarity to the reference image relative to the unedited source, together with Mask IoU to the source view.
Fidelity-only prompts provide limited gains because they do not specify that the source viewpoint should be retained, while asking the editor to generate a full 3D model substantially changes the input and produces a negative CLIP improvement.
Viewpoint instructions perform better overall; among them, ``\textit{Rotate the camera}'' directly requests another view of the same object without prescribing a specific azimuth or elevation.
As shown in Table~\ref{tab:prompt_ablation}, adding ``\textit{White background}'' gives the highest $\Delta$CLIP, a strong $\Delta$DINO, and high Mask IoU.
It also prevents the editor from introducing scene backgrounds that are absent from the white-background source renderings.
We therefore use ``\textit{Rotate the camera. White background.}'' for all experiments.

\begin{table}[t]
  \centering
  \caption{Prompt selection for Reinforced Editing on the Conceptual Design Dataset. All entries share the same editor and RE hyper-parameters ($N=12$, $N_e=9$, $w=4$); we report $\Delta$CLIP / $\Delta$DINO similarity to $x^{\texttt{ref}}$ relative to the unedited $x^{\texttt{src}}$, and Mask IoU with $x^{\texttt{src}}$. \textbf{Bold} row is our final choice.}
  \label{tab:prompt_ablation}
  \setlength{\tabcolsep}{4pt}
  \begin{tabular}{@{}lccc@{}}
    \toprule
    Prompt & $\Delta$CLIP $\uparrow$ & $\Delta$DINO $\uparrow$ & IoU $\uparrow$ \\
    \midrule
    \multicolumn{4}{@{}l}{\textit{Fidelity-only}} \\
    Super resolution of the input image           & +0.0209 & +0.0371 & 0.9398 \\
    Enrich the detail given the reference image   & +0.0171 & +0.0250 & 0.9469 \\
    Add the appearance of image 1                 & +0.0233 & +0.0346 & 0.9406 \\
    \midrule
    \multicolumn{4}{@{}l}{\textit{Viewpoint change}} \\
    Move the camera                                & +0.0277 & +0.0398 & 0.9406 \\
    Obtain another side view                       & +0.0250 & +0.0341 & 0.9330 \\
    Generate a novel view                          & +0.0161 & +0.0264 & 0.9437 \\
    Move the camera to a novel view                & +0.0268 & +0.0387 & 0.9408 \\
    \midrule
    \multicolumn{4}{@{}l}{\textit{Full 3D reinterpretation}} \\
    Generate a 3D model of the image               & $-0.0274$ & +0.0062 & 0.9455 \\
    \midrule
    \multicolumn{4}{@{}l}{\textit{``Rotate the camera'' variants}} \\
    Rotate the camera                              & +0.0314 & +0.0371 & 0.9402 \\
    Rotate the camera. Consistent Lighting         & +0.0204 & +0.0357 & 0.9417 \\
    Rotate the camera. Consistent visual style     & +0.0206 & +0.0310 & 0.9428 \\
    Rotate the camera. Consistent concept design   & +0.0265 & +0.0316 & 0.9422 \\
    \textbf{Rotate the camera. White background.}\ (Ours) & \textbf{+0.0339} & \textbf{+0.0387} & \textbf{0.9411} \\
    \bottomrule
  \end{tabular}
\end{table}

\textbf{Off-the-shelf editor baselines.}
For the feedback comparison in Sec.~4.1 of the main paper, we compare against the original Qwen-Image-Edit-2511 pipeline and NanoBanana Pro.
Because both editors expose a single visual context that concatenates the source view $x^{\texttt{src}}$ and the reference $x^{\texttt{ref}}$ side by side, the prompt must first assign roles to the two images before requesting the edit:
\begin{quote}
\itshape
\raggedright
``The first image is a 3D rendered view. The second\\
image is the reference. Edit the first image to match\\
the reference object's appearance. White background.''
\end{quote}
The prompt is a minimal adaptation of the effective instruction reported by Photo3D and is applied identically to both editors.
Only the editor backbone changes between runs; the released source code lists API-specific parameters, including the sampler, classifier-free guidance scale, and denoising steps.

\section{User Preference Study Protocols}
\label{sec:user_study_details}
We conduct a user preference study to complement the automatic metrics in the main paper.
The study assesses whether generated 3D assets preserve the input image's identity and improve visual fidelity across rendered views.

\textbf{Study setup.}
We use the same 100 held-out examples from the Conceptual Design Dataset as in the main evaluation.
For each example, participants are shown the input reference image and multi-view renderings generated by three methods: Pretrained (Trellis), Photo3D, and OREO.
With method names hidden, participants choose the result with the best overall quality.

\textbf{Evaluation criteria.}
Participants are instructed to consider three aspects jointly.
First, \textit{Fidelity} measures whether the generated asset contains realistic materials, sharp boundaries, and fine-grained texture details.
Second, \textit{Consistency} measures whether the rendered views remain coherent as a 3D object rather than showing view-dependent artifacts.
Third, \textit{Identity} measures whether the asset preserves the input image's semantic identity and distinctive details.

\textbf{Participants.}
The study includes 20 participants, including participants familiar with 3D reconstruction and computer graphics.
All participants evaluate the same 100 examples, and we aggregate their preferences over all responses.

\begin{table}[t]
  \centering
  \caption{User preferences on 100 held-out Conceptual Design Dataset examples, assessing reference fidelity, 3D consistency, and identity preservation across views.}
  \label{tab:user_preference}
  \begin{tabular}{@{}lc@{}}
    \toprule
    Method & Preference Ratio $\uparrow$ \\
    \midrule
    Pretrained (Trellis) & 29\% \\
    Photo3D & 33\% \\
    OREO (Ours) & \textbf{38\%} \\
    \bottomrule
  \end{tabular}
\end{table}

As shown in Table~\ref{tab:user_preference}, OREO receives the highest aggregate preference share, with 38\% of all responses.
With only aggregate preference ratios recorded, we report descriptive results without unsupported significance tests.

\section{Computational Overhead}
\label{sec:computational_overhead}
Each OREO training iteration consists of two costs: a generator update (rollout, render, and backward pass), and a call to the frozen 2D editor to produce the edited rendering $x^{\texttt{tgt}}$.
The generator update alone takes roughly 22~s per iteration.
If the 9-step editing call is executed sequentially inside the same iteration, it dominates the wall-clock cost, contributing about 81\% of the total per-iteration time and stretching one iteration to roughly 5$\times$ the generator-only cost.
\section{Limitations and Failure Cases}
\label{sec:limitations_failure}
OREO inherits its appearance guidance from the 2D editing prior, and thus may also inherit the editor's compositional or viewpoint biases in rare cases.
We discuss three representative failure modes on the Conceptual Design Dataset and a limitation for generators that decouple geometry and appearance.

\textbf{Viewpoint drift on face-like objects.}
For face-like objects or characters, the 2D editor may rotate the face toward the camera instead of preserving the intended 3D-facing direction, leading to view-dependent orientation inconsistency.
This drift arises from a mismatch between the 2D editor's canonical portrait prior and the 3D view-consistency requirement, so edited renderings may reorient a face toward the camera; repeated use of such targets can improve local facial details while causing inconsistent face orientation across views.

\textbf{Editor-inherited color and style bias.}
The dynamic noise update tightens alignment with the editor's appearance prior and can therefore inherit editor-specific color bias on unusual materials such as stone or metallic surfaces, shifting the intended hue while preserving geometry.
Stylized materials outside the editor's photographic training distribution may also undergo style shifts.

\textbf{Remaining artifacts on weakly constrained regions.}
Rarely observed regions such as occluded surfaces, thin appendages, and sharp material boundaries receive weaker per-view supervision and can retain small local artifacts on back views and along thin structures.
These regions are also under-constrained by the reference $x^{\texttt{ref}}$, which only depicts a single canonical viewpoint.

\textbf{Requirements on the generator architecture.}
OREO requires the optimized stage to provide a conditional flow-based latent model and a frozen pretrained counterpart for constructing the positive and negative predictions.
In Trellis, this requirement is satisfied by both stages: Stage 1 represents the sparse structure with a latent on a dense 3D grid, whereas Stage 2 generates sparse SLAT features at the active coordinates.
The contrastive formulation can therefore be applied to Stage 1, Stage 2, or both.
For generators that decouple geometry and appearance, however, the improvements brought by OREO may be less consistent, as the editing signal cannot jointly constrain both components.

These observations suggest future directions such as stronger multi-view consistency constraints, 3D-aware editing models, or uncertainty-aware filtering of edited renderings before they are used as 2D pseudo-targets for generator updates.

\bibliographystyle{splncs04}
\bibliography{main}

@String(CVPR  = {IEEE Conf. Comput. Vis. Pattern Recog.})

@String(ICCV  = {Int. Conf. Comput. Vis.})

@String(ECCV  = {Eur. Conf. Comput. Vis.})

@String(NeurIPS = {Adv. Neural Inform. Process. Syst.})

@String(ICML  = {Int. Conf. Mach. Learn.})

@String(ICLR  = {Int. Conf. Learn. Represent.})

@String(AAAI  = {AAAI})

@String(TOG   = {ACM Trans. Graph.})

@String(CVPR  = {CVPR})

@String(ICCV  = {ICCV})

@String(ECCV  = {ECCV})

@String(NeurIPS = {NeurIPS})

@String(ICML  = {ICML})

@String(ICLR  = {ICLR})

@String(TOG   = {ACM TOG})

@inproceedings{radford2021learning,
  title={Learning transferable visual models from natural language supervision},
  author={Radford, Alec and Kim, Jong Wook and Hallacy, Chris and Ramesh, Aditya and Goh, Gabriel and Agarwal, Sandhini and Sastry, Girish and Askell, Amanda and Mishkin, Pamela and Clark, Jack and others},
  booktitle={ICML},
  pages={8748--8763},
  year={2021},
  organization={PmLR}
}

@inproceedings{ma2025progressive,
  title={Progressive Rendering Distillation: Adapting Stable Diffusion for Instant Text-to-Mesh Generation without 3D Data},
  author={Ma, Zhiyuan and Liang, Xinyue and Wu, Rongyuan and Zhu, Xiangyu and Lei, Zhen and Zhang, Lei},
  booktitle={CVPR},
  pages={11036--11050},
  year={2025}
}

@inproceedings{xiang2025structured,
  title={Structured 3d latents for scalable and versatile 3d generation},
  author={Xiang, Jianfeng and Lv, Zelong and Xu, Sicheng and Deng, Yu and Wang, Ruicheng and Zhang, Bowen and Chen, Dong and Tong, Xin and Yang, Jiaolong},
  booktitle={CVPR},
  pages={21469--21480},
  year={2025}
}

@inproceedings{chen20253dtopia,
  title={3dtopia-xl: Scaling high-quality 3d asset generation via primitive diffusion},
  author={Chen, Zhaoxi and Tang, Jiaxiang and Dong, Yuhao and Cao, Ziang and Hong, Fangzhou and Lan, Yushi and Wang, Tengfei and Xie, Haozhe and Wu, Tong and Saito, Shunsuke and others},
  booktitle={CVPR},
  pages={26576--26586},
  year={2025}
}

@inproceedings{deitke2023objaverse,
  title={Objaverse: A universe of annotated 3d objects},
  author={Deitke, Matt and Schwenk, Dustin and Salvador, Jordi and Weihs, Luca and Michel, Oscar and VanderBilt, Eli and Schmidt, Ludwig and Ehsani, Kiana and Kembhavi, Aniruddha and Farhadi, Ali},
  booktitle={CVPR},
  pages={13142--13153},
  year={2023}
}

@article{zhao2025hunyuan3d,
  title={Hunyuan3d 2.0: Scaling diffusion models for high resolution textured 3d assets generation},
  author={Zhao, Zibo and Lai, Zeqiang and Lin, Qingxiang and Zhao, Yunfei and Liu, Haolin and Yang, Shuhui and Feng, Yifei and Yang, Mingxin and Zhang, Sheng and Yang, Xianghui and others},
  journal={arXiv preprint arXiv:2501.12202},
  year={2025}
}

@article{li2025step1x,
  title={Step1x-3d: Towards high-fidelity and controllable generation of textured 3d assets},
  author={Li, Weiyu and Zhang, Xuanyang and Sun, Zheng and Qi, Di and Li, Hao and Cheng, Wei and Cai, Weiwei and Wu, Shihao and Liu, Jiarui and Wang, Zihao and others},
  journal={arXiv preprint arXiv:2505.07747},
  year={2025}
}

@inproceedings{ye2025hi3dgen,
  title={{Hi3DGen}: High-fidelity {3D} Geometry Generation from Images via Normal Bridging},
  author={Ye, Chongjie and Wu, Yushuang and Lu, Ziteng and Chang, Jiahao and Guo, Xiaoyang and Zhou, Jiaqing and Zhao, Hao and Han, Xiaoguang},
  booktitle={ICCV},
  pages={25050--25061},
  year={2025}
}

@inproceedings{downs2022google,
  title={Google scanned objects: A high-quality dataset of 3d scanned household items},
  author={Downs, Laura and Francis, Anthony and Koenig, Nate and Kinman, Brandon and Hickman, Ryan and Reymann, Krista and McHugh, Thomas B and Vanhoucke, Vincent},
  booktitle={ICRA},
  pages={2553--2560},
  year={2022},
  organization={IEEE}
}

@inproceedings{yushi2025gaussiananything,
  title={{GaussianAnything}: Interactive Point Cloud Flow Matching for {3D} Object Generation},
  author={Lan, Yushi and Zhou, Shangchen and Lyu, Zhaoyang and Hong, Fangzhou and Yang, Shuai and Dai, Bo and Pan, Xingang and Loy, Chen Change},
  booktitle={ICLR},
  year={2025}
}

@article{guo2025hyper3d,
  title={Hyper3d: Efficient 3d representation via hybrid triplane and octree feature for enhanced 3d shape variational auto-encoders},
  author={Guo, Jingyu and Gao, Sensen and Bian, Jia-Wang and Sun, Wanhu and Zheng, Heliang and Jia, Rongfei and Gong, Mingming},
  journal={arXiv preprint arXiv:2503.10403},
  year={2025}
}

@article{zhang2025bang,
  title={BANG: Dividing 3D Assets via Generative Exploded Dynamics},
  author={Zhang, Longwen and Zhang, Qixuan and Jiang, Haoran and Bai, Yinuo and Yang, Wei and Xu, Lan and Yu, Jingyi},
  journal={ACM Transactions on Graphics (TOG)},
  volume={44},
  number={4},
  pages={1--21},
  year={2025},
  publisher={ACM New York, NY, USA}
}

@article{lin2025diffsplat,
  title={Diffsplat: Repurposing image diffusion models for scalable gaussian splat generation},
  author={Lin, Chenguo and Pan, Panwang and Yang, Bangbang and Li, Zeming and Mu, Yadong},
  journal={arXiv preprint arXiv:2501.16764},
  year={2025}
}

@inproceedings{liang2026photo3d,
  title={Photo3D: Advancing Photorealistic 3D Generation through Structure-Aligned Detail Enhancement},
  author={Liang, Xinyue and Ma, Zhiyuan and Sun, Lingchen and Guo, Yanjun and Zhang, Lei},
  booktitle={Proceedings of the IEEE/CVF Conference on Computer Vision and Pattern Recognition},
  pages={34237--34247},
  year={2026}
}

@article{simeoni2025dinov3,
  title={Dinov3},
  author={Sim{\'e}oni, Oriane and Vo, Huy V and Seitzer, Maximilian and Baldassarre, Federico and Oquab, Maxime and Jose, Cijo and Khalidov, Vasil and Szafraniec, Marc and Yi, Seungeun and Ramamonjisoa, Micha{\"e}l and others},
  journal={arXiv preprint arXiv:2508.10104},
  year={2025}
}

@inproceedings{lin2023magic3d,
  title={Magic3d: High-resolution text-to-3d content creation},
  author={Lin, Chen-Hsuan and Gao, Jun and Tang, Luming and Takikawa, Towaki and Zeng, Xiaohui and Huang, Xun and Kreis, Karsten and Fidler, Sanja and Liu, Ming-Yu and Lin, Tsung-Yi},
  booktitle={CVPR},
  pages={300--309},
  year={2023}
}

@inproceedings{wang2024prolificdreamer,
  title={{ProlificDreamer}: High-Fidelity and Diverse Text-to-{3D} Generation with Variational Score Distillation},
  author={Wang, Zhengyi and Lu, Cheng and Wang, Yikai and Bao, Fan and Li, Chongxuan and Su, Hang and Zhu, Jun},
  booktitle={NeurIPS},
  volume={36},
  year={2023}
}

@inproceedings{ho2021classifier,
  title={Classifier-Free Diffusion Guidance},
  author={Ho, Jonathan and Salimans, Tim},
  booktitle={NeurIPS Workshop},
  year={2021}
}

@article{hunyuan3d2025hunyuan3d,
  title={Hunyuan3D 2.1: From Images to High-Fidelity 3D Assets with Production-Ready PBR Material},
  author={Hunyuan3D, Team and Yang, Shuhui and Yang, Mingxin and Feng, Yifei and Huang, Xin and Zhang, Sheng and He, Zebin and Luo, Di and Liu, Haolin and Zhao, Yunfei and others},
  journal={arXiv preprint arXiv:2506.15442},
  year={2025}
}

@inproceedings{hertz2022prompt,
  title={Prompt-to-prompt image editing with cross attention control},
  author={Hertz, Amir and Mokady, Ron and Tenenbaum, Jay and Aberman, Kfir and Pritch, Yael and Cohen-Or, Daniel},
  booktitle={ICLR},
  year={2023}
}

@inproceedings{mokady2023null,
  title={Null-text inversion for editing real images using guided diffusion models},
  author={Mokady, Ron and Hertz, Amir and Aberman, Kfir and Pritch, Yael and Cohen-Or, Daniel},
  booktitle={CVPR},
  year={2023}
}

@inproceedings{brooks2023instructpix2pix,
  title={Instructpix2pix: Learning to follow image editing instructions},
  author={Brooks, Tim and Holynski, Aleksander and Efros, Alexei A},
  booktitle={CVPR},
  year={2023}
}

@inproceedings{couairon2024flowedit,
  title={{FlowEdit}: Inversion-Free Text-Based Editing Using Pre-Trained Flow Models},
  author={Kulikov, Vladimir and Kleiner, Matan and Huberman-Spiegelglas, Inbar and Michaeli, Tomer},
  booktitle={ICCV},
  pages={19721--19730},
  year={2025}
}

@inproceedings{zhang2024rfinversion,
  title={Semantic Image Inversion and Editing using Rectified Stochastic Differential Equations},
  author={Rout, Litu and Chen, Yujia and Ruiz, Nataniel and Caramanis, Constantine and Shakkottai, Sanjay and Chu, Wen-Sheng},
  booktitle={ICLR},
  year={2025}
}

@String(CVPR= {IEEE Conf. Comput. Vis. Pattern Recog.})

@String(ICCV= {Int. Conf. Comput. Vis.})

@String(ECCV= {Eur. Conf. Comput. Vis.})

@String(TOG= {ACM Trans. Graph.})

@String(ICLR = {Int. Conf. Learn. Represent.})

@String(AAAI = {AAAI})

@article{poole2022dreamfusion,
  title={Dreamfusion: Text-to-3d using 2d diffusion},
  author={Poole, Ben and Jain, Ajay and Barron, Jonathan T and Mildenhall, Ben},
  journal={arXiv preprint arXiv:2209.14988},
  year={2022}
}

@article{lai2025hunyuan3d2_5,
  title={Hunyuan3D 2.5: Towards High-Fidelity 3D Assets Generation with Ultimate Details},
  author={Lai, Zeqiang and Zhao, Yunfei and Liu, Haolin and Zhao, Zibo and Lin, Qingxiang and Shi, Huiwen and Yang, Xianghui and Yang, Mingxin and Yang, Shuhui and Feng, Yifei and others},
  journal={arXiv preprint arXiv:2506.16504},
  year={2025}
}

@article{li2025triposg,
  title={TripoSG: High-Fidelity 3D Shape Synthesis using Large-Scale Rectified Flow Models},
  author={Li, Yangguang and Zou, Zi-Xin and Liu, Zexiang and Wang, Dehu and Liang, Yuan and Yu, Zhipeng and Liu, Xingchao and Guo, Yuan-Chen and Liang, Ding and Ouyang, Wanli and others},
  journal={arXiv preprint arXiv:2502.06608},
  year={2025}
}

@article{yang2024hunyuan3d,
  title={Hunyuan3d-1.0: A unified framework for text-to-3d and image-to-3d generation},
  author={Yang, Xianghui and Shi, Huiwen and Zhang, Bowen and Yang, Fan and Wang, Jiacheng and Zhao, Hongxu and Liu, Xinhai and Wang, Xinzhou and Lin, Qingxiang and Yu, Jiaao and others},
  journal={arXiv preprint arXiv:2411.02293},
  year={2024}
}

@inproceedings{yin2024onestep,
    title={One-step diffusion with distribution matching distillation},
    author={Yin, Tianwei and Gharbi, Micha{\"e}l and Zhang, Richard and Shechtman, Eli and Durand, Fr{\'e}do and Freeman, William T and Park, Taesung},
    booktitle={2024 IEEE/CVF Conference on Computer Vision and Pattern Recognition (CVPR)},
    pages={6613--6623},
    year={2024},
    organization={IEEE}
  }

@article{sheynin2023emu,
  title={Emu edit: Precise image editing via recognition and generation tasks},
  author={Sheynin, Shelly and Polyak, Adam and Singer, Uriel and Kirstain, Yuval and Zohar, Amit and Ashual, Oron and Parikh, Devi and Taigman, Yaniv},
  journal={arXiv preprint arXiv:2311.10089},
  year={2023}
}

@article{zhang2023magicbrush,
  title={Magicbrush: A manually annotated dataset for instruction-guided image editing},
  author={Zhang, Kai and Mo, Lingbo and Chen, Wenhu and Sun, Huan and Su, Yu},
  journal={NeurIPS},
  volume={36},
  pages={31428--31449},
  year={2023}
}

@article{zhao2024ultraedit,
  title={Ultraedit: Instruction-based fine-grained image editing at scale},
  author={Zhao, Haozhe and Ma, Xiaojian and Chen, Liang and Si, Shuzheng and Wu, Rujie and An, Kaikai and Yu, Peiyu and Zhang, Minjia and Li, Qing and Chang, Baobao},
  journal={NeurIPS},
  volume={37},
  pages={3058--3093},
  year={2024}
}

@inproceedings{haque2023instruct,
  title={Instruct-nerf2nerf: Editing 3d scenes with instructions},
  author={Haque, Ayaan and Tancik, Matthew and Efros, Alexei A and Holynski, Aleksander and Kanazawa, Angjoo},
  booktitle={ICCV},
  pages={19740--19750},
  year={2023}
}

@inproceedings{zhuang2023dreameditor,
  title={Dreameditor: Text-driven 3d scene editing with neural fields},
  author={Zhuang, Jingyu and Wang, Chen and Lin, Liang and Liu, Lingjie and Li, Guanbin},
  booktitle={SIGGRAPH Asia 2023 conference papers},
  pages={1--10},
  year={2023}
}

@inproceedings{wang2024gaussianeditor,
  title={Gaussianeditor: Editing 3d gaussians delicately with text instructions},
  author={Wang, Junjie and Fang, Jiemin and Zhang, Xiaopeng and Xie, Lingxi and Tian, Qi},
  booktitle={CVPR},
  pages={20902--20911},
  year={2024}
}

@inproceedings{koh2026diffusion,
  title={Diffusion Feature Field for Text-based {3D} Editing with Gaussian Splatting},
  author={Koh, Eunseo and Hyun, Sangeek and Lee, MinKyu and Chung, Jiwoo and Seo, Kangmin and Heo, Jae-Pil},
  booktitle={NeurIPS},
  volume={38},
  pages={159--178},
  year={2025}
}

@inproceedings{yenphraphai2024image,
  title={Image sculpting: Precise object editing with 3d geometry control},
  author={Yenphraphai, Jiraphon and Pan, Xichen and Liu, Sainan and Panozzo, Daniele and Xie, Saining},
  booktitle={CVPR},
  pages={4241--4251},
  year={2024}
}

@article{chen2024mvdrag3d,
  title={Mvdrag3d: Drag-based creative 3d editing via multi-view generation-reconstruction priors},
  author={Chen, Honghua and Lan, Yushi and Chen, Yongwei and Zhou, Yifan and Pan, Xingang},
  journal={arXiv preprint arXiv:2410.16272},
  year={2024}
}

@inproceedings{chen2024geodiffusion,
  title={Geodiffusion: Text-prompted geometric control for object detection data generation},
  author={Chen, Kai and Xie, Enze and Chen, Zhe and Wang, Yibo and Hong, Lanqing and Li, Zhenguo and Yeung, Dit-Yan},
  booktitle={ICLR},
  volume={2024},
  pages={8979--9001},
  year={2024}
}

@inproceedings{xie2026dnaedit,
  title={{DNAEdit}: Direct Noise Alignment for Text-Guided Rectified Flow Editing},
  author={Xie, Chenxi and Li, Minghan and Li, Shuai and Wu, Yuhui and Yi, Qiaosi and Zhang, Lei},
  booktitle={NeurIPS},
  volume={38},
  pages={124456--124474},
  year={2025}
}

@article{qin2020u2,
  title={U2-Net: Going deeper with nested U-structure for salient object detection},
  author={Qin, Xuebin and Zhang, Zichen and Huang, Chenyang and Dehghan, Masood and Zaiane, Osmar R and Jagersand, Martin},
  journal={Pattern recognition},
  volume={106},
  pages={107404},
  year={2020},
  publisher={Elsevier}
}

@inproceedings{yu2026self,
  title={Self-evaluation unlocks any-step text-to-image generation},
  author={Yu, Xin and Qi, Xiaojuan and Li, Zhengqi and Zhang, Kai and Zhang, Richard and Lin, Zhe and Shechtman, Eli and Wang, Tianyu and Nitzan, Yotam},
  booktitle={CVPR},
  pages={7816--7826},
  year={2026}
}

@article{yang20264dvd,
  title={4DVD: cascaded dense-view video diffusion model for high-quality 4D content generation},
  author={Yang, Shuzhou and Cun, Xiaodong and Li, Xiaoyu and Li, Yaowei and Zhang, Jian},
  journal={International Journal of Computer Vision},
  volume={134},
  number={5},
  pages={233},
  year={2026},
  publisher={Springer}
}

@article{yang2025hybrid,
  title={Hybrid Fourier score distillation for efficient one image to 3D object generation},
  author={Yang, Shuzhou and Wang, Yu and Li, Haijie and Meng, Jiarui and Wu, Yanmin and Meng, Xiandong and Zhang, Jian},
  journal={Visual Intelligence},
  volume={3},
  number={1},
  pages={17},
  year={2025},
  publisher={Springer}
}

@inproceedings{
yang2026gencompositor,
title={GenCompositor: Generative Video Compositing with Diffusion Transformer},
author={Shuzhou Yang and Xiaoyu Li and Xiaodong Cun and Guangzhi Wang and Lingen Li and Ying Shan and Jian Zhang},
booktitle={The Fourteenth International Conference on Learning Representations},
year={2026},
url={https://openreview.net/forum?id=ynim5u2N4i}
}

@inproceedings{liang2026aligncvc,
  title={Aligncvc: Aligning cross-view consistency for single-image-to-3d generation},
  author={Liang, Xinyue and Ma, Zhiyuan and Sun, Lingchen and Guo, Yanjun and Zhang, Lei},
  booktitle={AAAI},
  volume={40},
  pages={6889--6897},
  year={2026}
}

@inproceedings{chen2026omni,
  title={Omni-3DEdit: Generalized Versatile 3D Editing in One-Pass},
  author={Chen, Liyi and Wang, Pengfei and Zhang, Guowen and Ma, Zhiyuan and Zhang, Lei},
  booktitle={CVPR},
  pages={12640--12650},
  year={2026}
}

@inproceedings{liu2025mvboost,
  title={Mvboost: Boost 3d reconstruction with multi-view refinement},
  author={Liu, Xiangyu and Zhang, Xiaomei and Ma, Zhiyuan and Zhu, Xiangyu and Lei, Zhen},
  booktitle={CVPR},
  pages={21664--21673},
  year={2025}
}

@article{wang2026one2scene,
  title={One2scene: Geometric consistent explorable 3d scene generation from a single image},
  author={Wang, Pengfei and Chen, Liyi and Ma, Zhiyuan and Guo, Yanjun and Zhang, Guowen and Zhang, Lei},
  journal={arXiv preprint arXiv:2602.19766},
  year={2026}
}

@article{guo2026memorize,
  title={Memorize When Needed: Decoupled Memory Control for Spatially Consistent Long-Horizon Video Generation},
  author={Guo, Yanjun and Zhang, Zhengqiang and Wang, Pengfei and Liang, Xinyue and Ma, Zhiyuan and Zhang, Lei},
  journal={arXiv preprint arXiv:2604.18215},
  year={2026}
}

@inproceedings{ma2024scaledreamer,
  title={Scaledreamer: Scalable text-to-3d synthesis with asynchronous score distillation},
  author={Ma, Zhiyuan and Wei, Yuxiang and Zhang, Yabin and Zhu, Xiangyu and Lei, Zhen and Zhang, Lei},
  booktitle={ECCV},
  pages={1--19},
  year={2024},
  organization={Springer}
}

@article{wu2025qwen,
  title={Qwen-image technical report},
  author={Wu, Chenfei and Li, Jiahao and Zhou, Jingren and Lin, Junyang and Gao, Kaiyuan and Yan, Kun and Yin, Sheng-ming and Bai, Shuai and Xu, Xiao and Chen, Yilei and others},
  journal={arXiv preprint arXiv:2508.02324},
  year={2025}
}

\end{document}